\documentclass{article}

\usepackage[eandd, final]{neurips_2026}

\usepackage[utf8]{inputenc} 
\usepackage[T1]{fontenc}    
\usepackage{hyperref}       
\usepackage{url}            
\usepackage{booktabs}       
\usepackage{amsfonts}       
\usepackage{nicefrac}       
\usepackage{microtype}      
\usepackage{xcolor}         

\usepackage{graphicx}
\usepackage{subcaption}

\usepackage{amsmath}
\usepackage{amssymb}
\usepackage{mathtools}
\usepackage{amsthm}

\newcommand{\structc}{\textsc{Struct-CIFAR100-20}}
\newcommand{\seqc}{\textsc{Seq-CIFAR100-10}}
\newcommand{\seqct}{\textsc{Seq-CIFAR100-20}}
\newcommand{\seqm}{\textsc{Seq-MNIST-5}}
\newcommand{\rotm}{\textsc{Rot-MNIST-20}}

\title{Re-evaluating Continual Learning\\with Few-Shot Adaptation}

\author{%
  Amogh Inamdar \\
  Department of Computer Science\\
  Columbia University\\
  \texttt{amogh.inamdar@columbia.edu} \\
  \And
  Matthew So \\
  Department of Computer Science\\
  Columbia University\\
  \texttt{ms5513@columbia.edu} \\
  \And
  Vici Milenia \\
  Department of Computer Science\\
  Columbia University\\
  \texttt{vim2111@columbia.edu} \\
  \And
  Richard Zemel\\
  Department of Computer Science\\
  Columbia University \\
  \texttt{zemel@cs.columbia.edu} \\
}

\begin{document}

\maketitle

\begin{abstract}
  Continual learning methods aim to maximize the stability and plasticity of machine learning models that are trained on a sequence of tasks. The standard measure of stability (i.e., forgetting) is the 0-shot performance of a model on previously learned tasks, and plasticity, the performance on the most recently learned task. However, 0-shot evaluation does not fully measure a model or method's ability to retain learned information or adapt quickly to new information, as it requires perfect recall across multiple tasks. In this paper, we propose \emph{few-shot evaluation} as a more comprehensive assessment of the stability and plasticity of a continual learning system. We conduct a fine-grained assessment on task sequences for continual image classification and find that this paradigm produces novel insights into the performance of popular continual learning strategies. Through few-shot evaluation with a novel metric---\emph{per-shot plasticity}---we show that adding `foresight' to continual learning methods via the meta-learning of a short sequence of future tasks induces \emph{learning-to-learn} behavior over the task sequence.
\end{abstract}

\section{Introduction}\label{sec:intro}

The \emph{stability-plasticity dilemma} of continual learning (CL) arises from the difficulty in retaining performance on prior tasks (`stability') while also maximizing performance on the current task (`plasticity') when training a machine learning (ML) model on a sequence of tasks. As ML models grow both increasingly capable and increasingly expensive to train, the need for continuous and incremental model updates has brought this dilemma to the forefront of ML research. At the time of writing, the standard measure of the \emph{stability} of a model as its resistance to \emph{catastrophic forgetting} \citep{mccloskey1989catastrophic, french1999catastrophic, goodfellow2015empirical}, or the irrecoverable loss of performance on previously learned tasks when a model learns a new task. \emph{Plasticity}, which represents how well a model can learn new information, is typically measured by the performance of a model on the task that it was most recently optimized to learn \cite{dohare2024maintainingplasticitydeepcontinual}. 

However, we find that this evaluation paradigm does not capture multiple key characteristics of continually trained models. Firstly, while a CL model may not show a high recall on previously learned tasks after optimizing on a new task, this `forgetting' is not necessarily catastrophic. Natural learners (like humans) often require warm-up via \emph{cued recall} to switch context and perform well on previously learned tasks \cite{tulving1973encoding}. 
The rapid re-acquisition of seemingly forgotten concepts, known as the \emph{relearning effect} \citep{ebbinghaus1885gedachtnis}, is a well-studied phenomenon in humans \cite{murre2015replication} and other animals \citep{napier1992rapid,ricker1996reacquisition}.
Similarly, a stable CL model may efficiently regain past performance---an aspect that is not measured by existing metrics for forgetting or backward transfer. Evaluating after \emph{adapting with a few task examples} may recover past information that is not completely inaccessible, but simply has a low probability of surfacing due to the most recent optimization. 

Secondly, while the mitigation of catastrophic forgetting (stability) is a well-studied problem in continual learning, plasticity is often treated as an afterthought, and measured only as the ability to learn a new task through direct optimization. In contrast, plasticity in learning systems can \emph{improve} as learning progresses, via, e.g., the learning of shared task structure \citep{ABRAHAM200573, benfenati2007synaptic}. A related metric that is occasionally studied in CL is 0-shot forward transfer \cite{lopez2017gradient}, which measures the average performance on tasks that have not yet been learned. However, this metric is trivially zero when new tasks do not share the same outputs (e.g. class labels) as learned tasks, and only measures the ability to generalize \emph{across} tasks in shared domains. An ideal metric for plasticity should also measure \emph{within-task adaptation}---how efficiently a model adapts to a new task, i.e., its relative improvement with limited information or training. This definition is better aligned with natural learning, and enables plasticity to be measured and aggregated over tasks with differing inputs, outputs, and relative difficulty.

In this work, we propose a novel paradigm for evaluation in continual learning that addresses both of these issues: \emph{continuous evaluation with few-shot adaptation}. Modern ML models, particularly for vision and language, exhibit an impressive \emph{few-shot learning} ability to perform tasks with very little new information and computation compared to full-fledged task training \cite{lake2011one, brown2020language}. We train neural networks on task sequences for vision classification and conduct a large-scale evaluation of each checkpoint on every learning task with few-shot adaptation, finding surprising results with popular CL methods. We introduce a novel metric---\emph{per-shot plasticity}---that measures the rapidity of adaptation of a model to a task, providing additional insight into the dynamics of continual learning. We also develop a simple but novel method that meta-initializes the learner on \emph{future samples} prior to each task, optimizing for learning-to-learn ability without sacrificing fine-tuning or backward performance.

\begin{figure}[ht]
\centering
\centerline{\includegraphics[width=\linewidth]{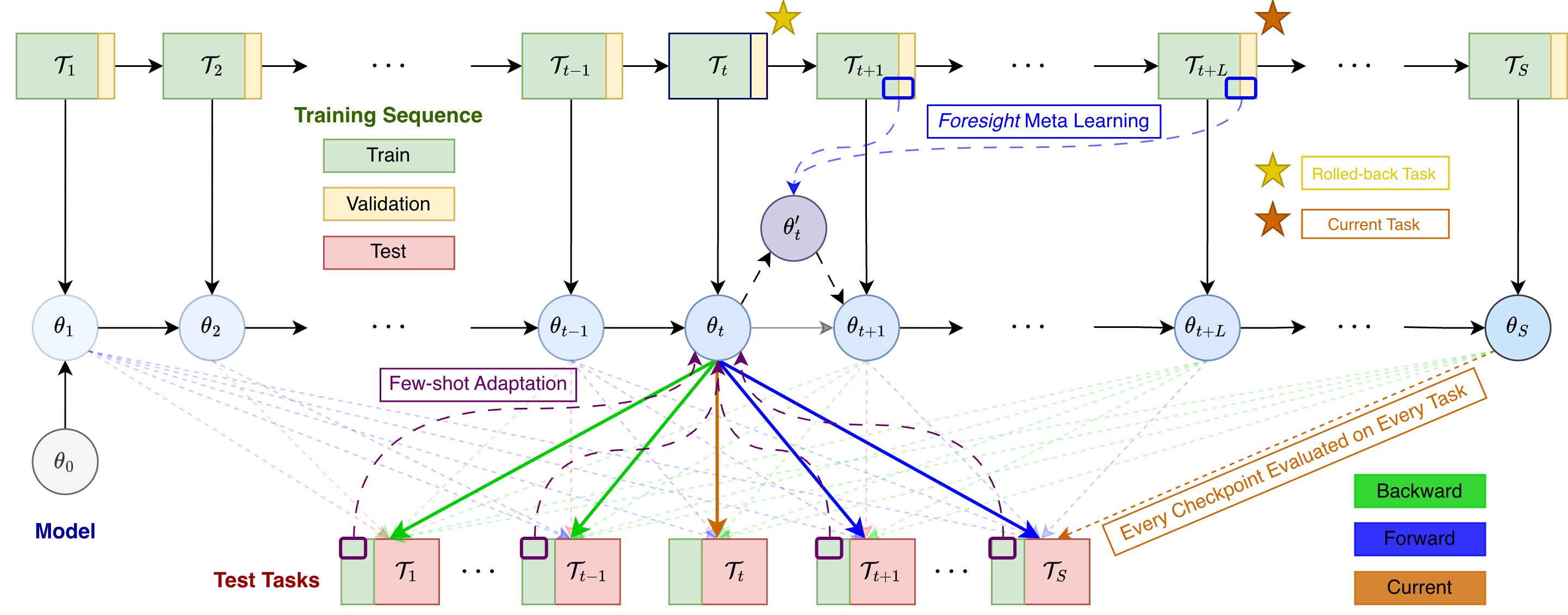}}
\caption{
  Few-shot evaluation and adaptation in continual learning. A model $\theta_0$ is trained on a task sequence $\mathcal{T}_1\rightarrow\cdots\rightarrow\mathcal{T}_S$ and evaluated on every task at each post-task checkpoint. We propose evaluation with \emph{few-shot adaptation} to the task data (\textcolor{violet}{violet}), and show surprising results with existing methods. We also develop a \emph{foresight meta-learning} algorithm (\textcolor{blue}{blue}) that rolls back the model at task $t+L$ to task $t$ and leverages the intermediate sequence to improve few-shot performance, and a novel metric for measuring plasticity and learning-to-learn ability as rapid adaptation (Section \ref{sec:sauce}).
}
\label{fig:intro-fig}
\vspace{-0.5cm}
\end{figure}
\section{Related Work}\label{sec:related}

\subsection{Continual Learning}

With the rise of powerful multi-task deep learning models, continual learning---the training and evaluation of models on a \emph{sequence of tasks}---is an increasingly popular paradigm in machine learning research. Each task in the sequence may add a new set of elements to existing objectives (\emph{class-incremental}), switch to an entirely different set of objectives (\emph{task-incremental}), or shift the training data distribution (\emph{domain-incremental} CL) \cite{wang2024comprehensive, bidaki2025online}.
Existing methods for continual learning focus primarily on mitigating \emph{catastrophic forgetting} \cite{mccloskey1989catastrophic, french1999catastrophic}. Of these, \emph{replay}-based strategies that focus on storing and re-training on past examples \cite{lin1992self, bang2021rainbow, aljundi2019onlinecontinuallearningmaximally, liu2020mnemonics} are the most popular \cite{bidaki2025online}. Other strategies focus on \emph{distilling} previous model states \cite{buzzega2020darkexperiencegeneralcontinual}, \emph{regularizing} the model to prevent information loss \cite{kirkpatrick2017overcoming, lopez2017gradient, AGEM}, or modifying the model \emph{architecture} as new tasks are learned \cite{li2017learningforgetting, ge2025dynamic}. We present a more extensive review in Appendix \ref{apx:more-related}.

While some recent work has shifted focus to the continual pre-training of massive-scale foundation models \citep{ibrahim2024simple, li2025tic}, the scale and high redundancy \cite{wang2023too} of pretraining data limit the ability to conduct robust experimentation, and likely also the effectiveness of curricula. Image classification remains the primary domain for studies on continual learning \cite{bidaki2025online}, and we focus on studying the continual image classification setting in this work. 

\subsection{Evaluation in Continual Learning}

Recent work has emphasized the importance of \emph{continuous evaluation} over the task sequence \cite{de2022continual} to fully characterize the effects of continual learning. Rather than a single metric summarizing a complex setting, the standard practice is to measure a model's `stability' on previously learned tasks \cite{french1999catastrophic}, and its `plasticity' on the most recently learned task \cite{dohare2024maintainingplasticitydeepcontinual}. 
Formally, let $\theta_0$ be an ML model trained with method $\Omega$ on a task sequence $\mathcal{T}_1\rightarrow\mathcal{T}_2\rightarrow\cdots\rightarrow\mathcal{T}_S$ to produce models $\theta_1,\theta_2,\ldots,\theta_S$ respectively. A common metric for stability is the \emph{Backward Transfer/Average Forgetting} $\mathcal{F}$, measured at task $\mathcal{T}_t$ as the average loss in performance of subsequent model states $\theta_{>t}$ \cite{lopez2017gradient}.
\begin{equation}\label{eqn:forgetting}
    \mathcal{F}(\Omega, \mathcal{T}_t) = \frac{1}{S-t}\sum_{j=t+1}^S \mathrm{A}(\theta_t^\Omega, \mathcal{T}_t) - \mathrm{A}(\theta_j^\Omega, \mathcal{T}_t)
\end{equation}
where $\mathrm{A}$ is a measure of performance like test-set accuracy. Stability is defined as $\sigma^{CL}=-\mathcal{F}$, as a stable model has a lower forgetting score. The standard measure of plasticity $\phi^\mathrm{CL}$ is the \emph{Average Task Accuracy}, which measures the performance on a task immediately after optimization.
\begin{equation}\label{eqn:cl-plasticity}
    \phi^\mathrm{CL}(\Omega, \mathcal{T}_t) = \frac{1}{S}\sum_{i=1}^S \mathrm{A}(\theta_i^\Omega, \mathcal{T}_i)
\end{equation}
However, this notion of plasticity is \emph{reactive}---it only ensures that stable methods retain the ability to learn via gradient descent.. As we subsequently demonstrate, this metric fails to capture the difference in adaptation ability between models trained with various CL methods---with a strong learner, $\phi^{CL}$ is high irrespective of the learning method.

While the goal of learning that transfers to future tasks is well-represented in the CL literature, we observe that most methods do not evaluate or optimize for \emph{forward transfer}. Forward transfer is predominantly measured by the 0-shot accuracy on new tasks \cite{lopez2017gradient}, which is trivially zero in settings where tasks do not share any common outputs or structure. In contrast, research on \emph{few-shot learning} explicitly studies the phenomenon of rapid adaptation to new tasks. For example, vision models can learn to categorize images of previously unseen classes by fine-tuning on only a handful of images (`shots') \cite{vinyals2016matching, snell2017prototypical}, and LLMs can use \emph{in-context learning} \cite{brown2020language, dong2024surveyincontextlearning} to perform new tasks without optimization by including instructions and examples in their task prompt.
To the best of our knowledge, the only exploration of few-shot transfer is the Learning Curve Area \cite{AGEM}, which measures the area under the curve as accuracy increases with $k$-shot adaptation.
\begin{equation}\label{eqn:lca}
    \mathrm{LCA}_K = \frac{1}{k+1} \int_0^K Z_k dk = \frac{1}{K+1}\sum_{k=0}^K Z_k
\end{equation}
where $Z_k$ is the average $k$-shot performance on a set of evaluation tasks. However, the LCA is not a true measure of `area': in effect, it measures the average performance over 0--$K$ shots. Thus, the LCA does not measure the \emph{rapidity of adaptation} over $K$ examples of a task. In this work, we define the \emph{per-shot plasticity} as the Scaled Area under the performance-Adaptation CurvE (SAUCE), which measures the improvement in performance with added few-shot examples. This metric captures the cumulative error in early adaptation, aligning with natural definitions of plasticity \citep{ABRAHAM200573, benfenati2007synaptic}.

\subsection{Meta-learning and continual learning}\label{related:meta-learn}

Ideally, a model that repeatedly learns tasks should not only get better at each task, but also improve its process of learning over time, particularly when there is an underlying cross-task structure. Put another way, the few-shot adaptation ability of a CL model should improve as it progresses through the task sequence. This objective of \emph{learning-to-learn} from a set of tasks is the focus of the field of meta learning \cite{hospedales2020metalearningneuralnetworkssurvey}. We provide an overview of meta learning in Appendix \ref{apx:meta-learning-overview}. Both CL and meta learning study variants of multi-task learning, but with the added constraint of a fixed sequence of tasks in CL. Existing works on continual meta-learning (meta-CL) can be divided into two strategies. The first strategy modifies the \emph{architecture} of the learner using meta-information across the task sequence \citep{beaulieu2020learning, javed2019meta}. The second strategy uses a \emph{meta-optimizer} to leverage cross-task information while updating the model on each task \cite{riemer2019learninglearnforgettingmaximizing, pmlr-v97-finn19a, gupta2020lamamllookaheadmetalearning, wu2024meta}. We summarize these methods in Appendix \ref{apx:meta-learning-overview}.

However, we observe multiple shortcomings in existing methods for meta-CL. First, many works only evaluate on short sequences (<10 tasks), and methods often do not scale well to long sequences. Second, these methods focus only on preventing forgetting, obtaining all meta-information from a buffer of past task examples \emph{that were already learned} to update the model. Finally, to our knowledge, no meta-CL method \emph{optimizes for few-shot performance} (either forward or backward), despite this being a prevalent goal in meta-learning studies. In this work, we develop a \emph{foresight} meta-CL method that explicitly optimizes for few-shot performance by meta learning on new examples.

\section{Stability and Plasticity in the Few-Shot Paradigm}

We now present a comprehensive comparison of 0- and few-shot evaluation in visual continual learning. We first study the differences between 0-shot and few-shot \emph{stability} and show that `catastrophic' forgetting is efficiently recoverable in certain settings (Section \ref{sec:few-shot-stability}). We then examine the \emph{plasticity} of CL methods through the lens of few-shot evaluation, and find differences that do not emerge when evaluated 0-shot (Section \ref{sec:few-shot-plasticity}). Observing minimal \emph{learning-to-learn} behavior with existing CL methods, we develop a foresight meta-CL method
(Section \ref{sec:foresight-meta}) and a novel metric to study the \emph{learning-to-learn} ability of CL models (Section \ref{sec:sauce}).

\subsection{Datasets and Methods}\label{sec:datasets}

We evaluate methods representing key strategies in continual learning on task sequences constructed from the most popular datasets for CL \citep{bidaki2025online}. {\seqm} splits the MNIST handwritten digit dataset \citep{lecun2010mnist} into five binary classification tasks (0/1, 2/3, etc.), producing a \emph{2-way 5-task} sequence. Similarly, {\seqc} and {\seqct} split the CIFAR100 dataset \cite{Krizhevsky09learningmultiple} into \emph{10-way 10-task} and \emph{5-way 20-task} sequences, with each task consisting of random groupings of classes. However, every five CIFAR100 classes correspond to one superclass, and the random groupings thus contain overlapping concepts between tasks. To study a setting with higher task independence, we construct {\structc}, a \emph{5-way 20-task} split of CIFAR100 along superclass labels \citep{zhang2024integrating}. We also construct {\rotm}, a domain-incremental \emph{10-way 20-task} sequence that progressively increases the rotation of each MNIST image. The first task is the standard 10-class MNIST dataset. Subsequently, we rotate each image by $20^\circ$ with a random jitter of up to $5^\circ$ producing a smooth variance in the input characteristics for each of the 10 output classes. This results in an intentionally `messy' scenario meant to represent more dynamic task interaction, with the first two and last two tasks being similar, and some input noise as the `6' and `9' images flip. We note that this version is a \emph{hard sequence} to learn: La-MAML \citep{gupta2020lamamllookaheadmetalearning} studies a 4-task variant of this sequence, which is too short to study forward transfer. 

We use a ResNet-18 \cite{he2016deep} backbone in each setting except for {\seqm}, where we use an MLP with 16 hidden units to prevent memorization. We evaluate the following methods: ER \citep{lin1992self} (replay), DER++ \citep{buzzega2020darkexperiencegeneralcontinual} (distillation), EWC \citep{kirkpatrick2017overcoming} and AGEM \citep{AGEM} (regularization), MER \cite{riemer2019learninglearnforgettingmaximizing} (meta-CL), and SGD, a baseline that simply trains the model on all data as encountered. For a fair comparison, we do not include methods that modify architectures (e.g., add parameters). We present a detailed description of our training and evaluation setup in Appendix \ref{apx:details}.

\subsection{Few-Shot Stability Highlights Non-Catastrophic Forgetting}\label{sec:few-shot-stability}

\begin{figure}
\centering
\begin{subfigure}{0.35\textwidth}
        \includegraphics[height=3cm]{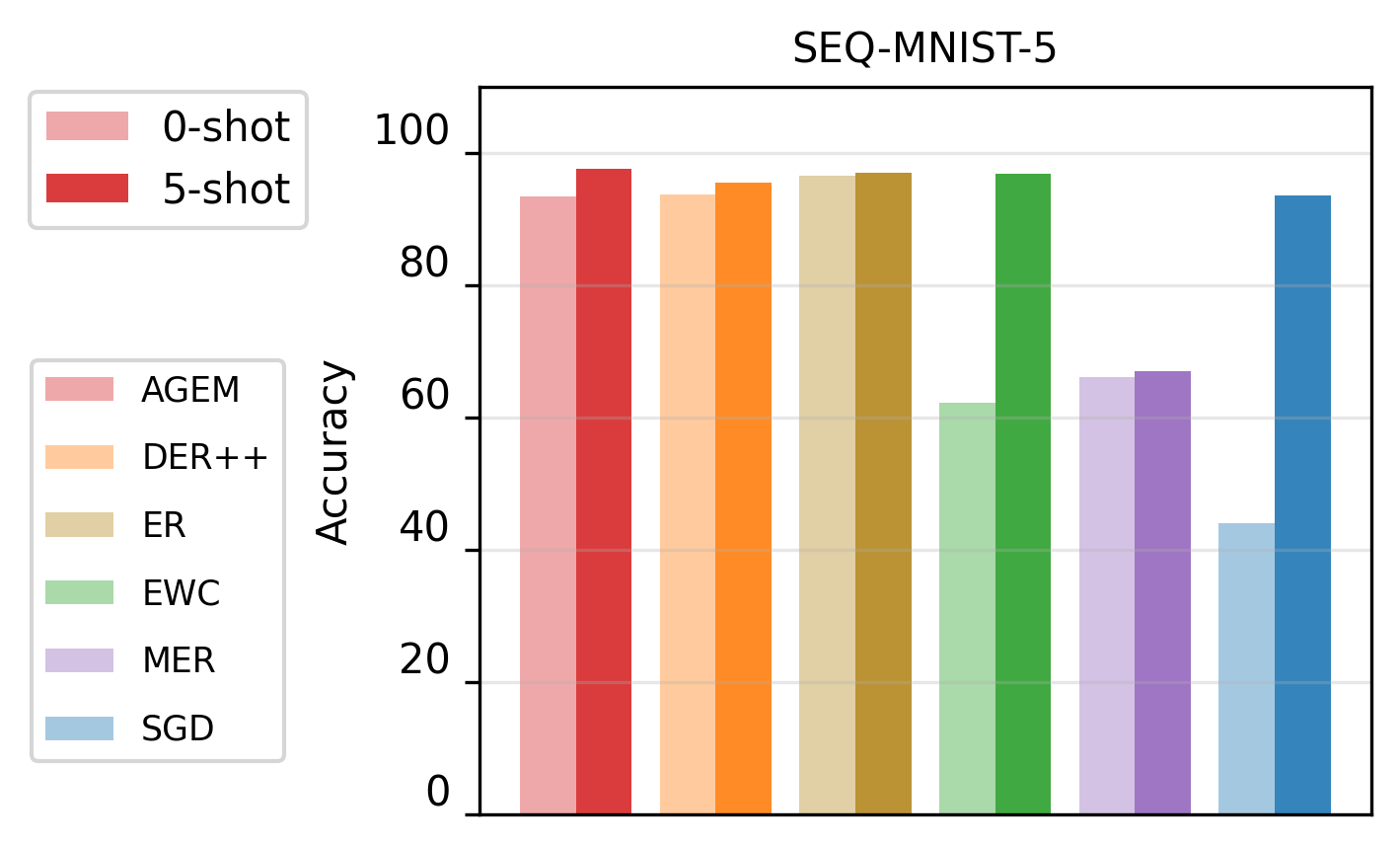}
    \label{fig:stability-mnist}
\end{subfigure}
\begin{subfigure}{0.21\textwidth}
        \includegraphics[height=3cm]{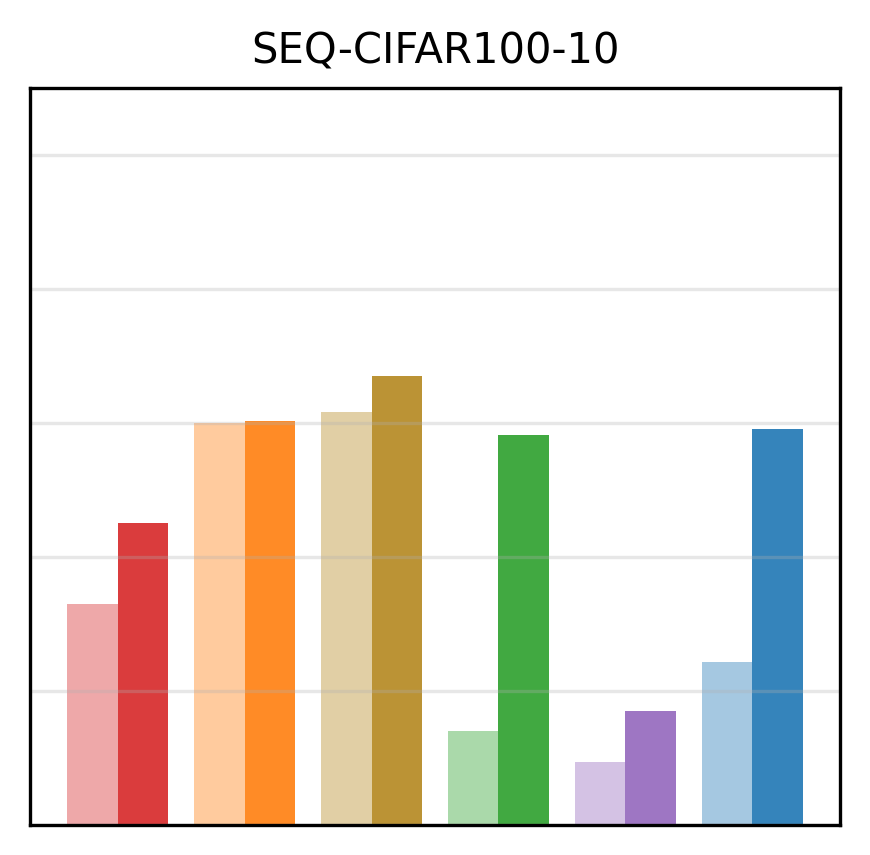}
    \label{fig:stability-seq-cifar100}
\end{subfigure}
\begin{subfigure}{0.21\textwidth}
        \includegraphics[height=3cm]{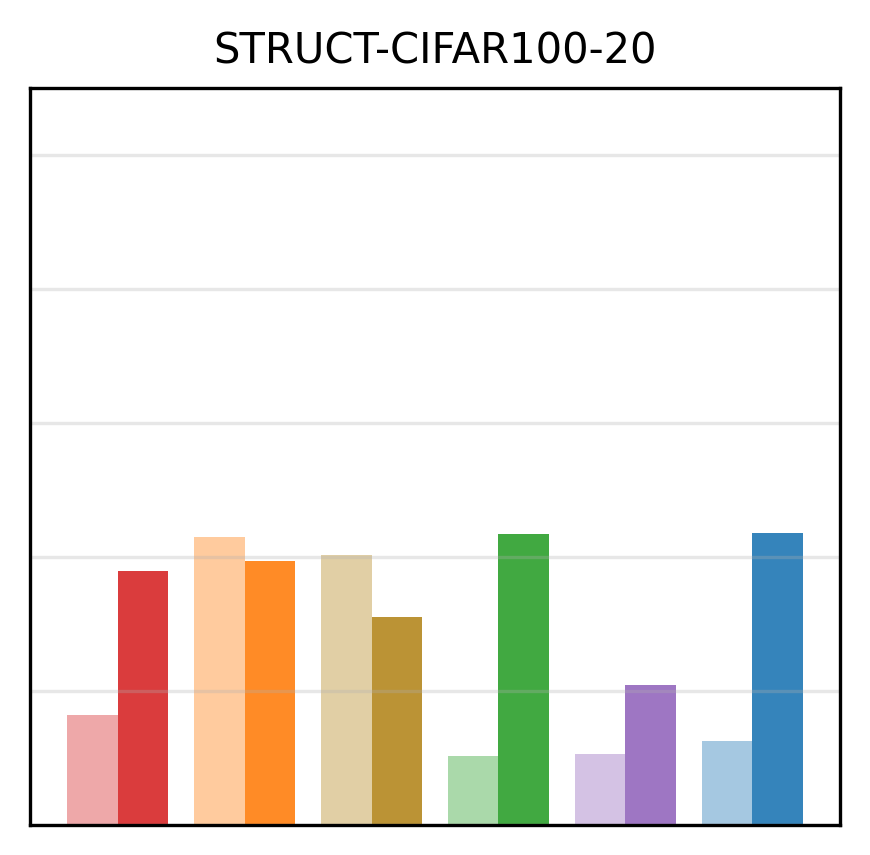}
    \label{fig:stability-struct-cifar100}
\end{subfigure}
\begin{subfigure}{0.21\textwidth}
        \includegraphics[height=3cm]{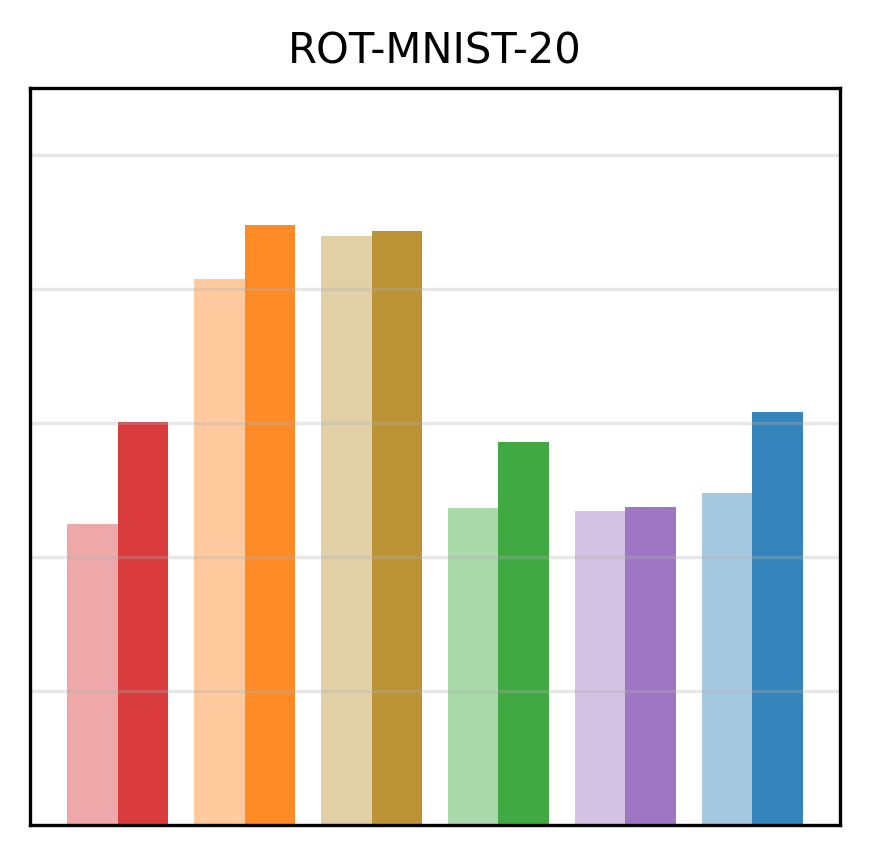}
    \label{fig:stability-rot-mnist}
\end{subfigure}
\caption{0-shot vs. 5-shot accuracy on \emph{backward} tasks in 4 common CL sequences. With 0-shot evaluation, replay-based methods almost match prior accuracy, but other methods show significant forgetting, especially the non-stability-preserving SGD baseline. However, 5-shot evaluation tells a different story---performance recovers drastically and nearly matches the prior best in every setting.}
\label{fig:stable-res-1}
\end{figure}

We evaluate \emph{every checkpoint on all past and future tasks after adapting with $k=0,1,2,5,10$ task examples}. Here, $k=0$ corresponds to the typical zero-shot evaluation paradigm. We emphasize the completeness of our evaluation---for a 20-task sequence and 5 values of $k$, this requires $S^2\times 5=2000$ test-set forward passes. We first evaluate the \emph{stability} of each method as the average accuracy on \emph{backward} tasks $\mathcal{T}_{<t}$. In the 0-shot setting, methods that explicitly replay task examples (ER \citep{lin1992self}) or model outputs (DER++ \citep{buzzega2020darkexperiencegeneralcontinual}) during training vastly outperform the fine-tuning baseline (SGD) on each dataset, with regularization-based methods also retaining strong performance. The performance of SGD is often close to random, indicating catastrophic forgetting. However, the results at $k=5$ tell a different story. SGD recovers a significant amount of performance, even approaching the accuracy of methods that use and store and replay samples during training. This increase in performance with few-shot adaptation on previously learned tasks is significantly higher than any corresponding increase in tasks that have not yet been learned, confirming that this is indeed the \emph{rapid reacquisition} \cite{jirenhed2007acquisition, bouton2004context} of prior knowledge. Thus, few-shot test-time replay appears to be nearly as effective as training-time replay, while reducing the inductive bias and incurring a much smaller overhead cost---samples only have to be replayed once at evaluation time. We present examples of per-checkpoint accuracy (Figures \ref{fig:accuracy-seq-c100}, \ref{fig:accuracy-smooth-mnist}) and a summary of metrics $k$-values (Figure \ref{fig:mnist-few-shot-full}) in Appendix \ref{apx:more-results}).

While Figure \ref{fig:stable-res-1} already provides novel information about stability, understanding the full picture of continual learning requires a more fine-grained analysis than an aggregation over checkpoints \citep{de2022continual}. This enables the measurement of the \emph{learning-to-learn} ability of a model as its rate of performance improvement over time. In the top row of Figure \ref{fig:improvement-res-1}, we present the progression over the learning sequence of the average 10-shot performance on previously learned tasks of a checkpoint $\theta_t$. We focus our evaluation on our 20-task sequences to better observe long-term effects. Every method (except MER) appears to maintain stability, showing \emph{no deterioration in capacity} at these sequence lengths. However, there is no significant improvement in stability over time for any methods.

\subsection{Few-Shot Forward Transfer Differentiates Method Plasticity}\label{sec:few-shot-plasticity}

\begin{figure}
\centering
\begin{subfigure}{0.35\textwidth}
        \includegraphics[height=2.8cm]{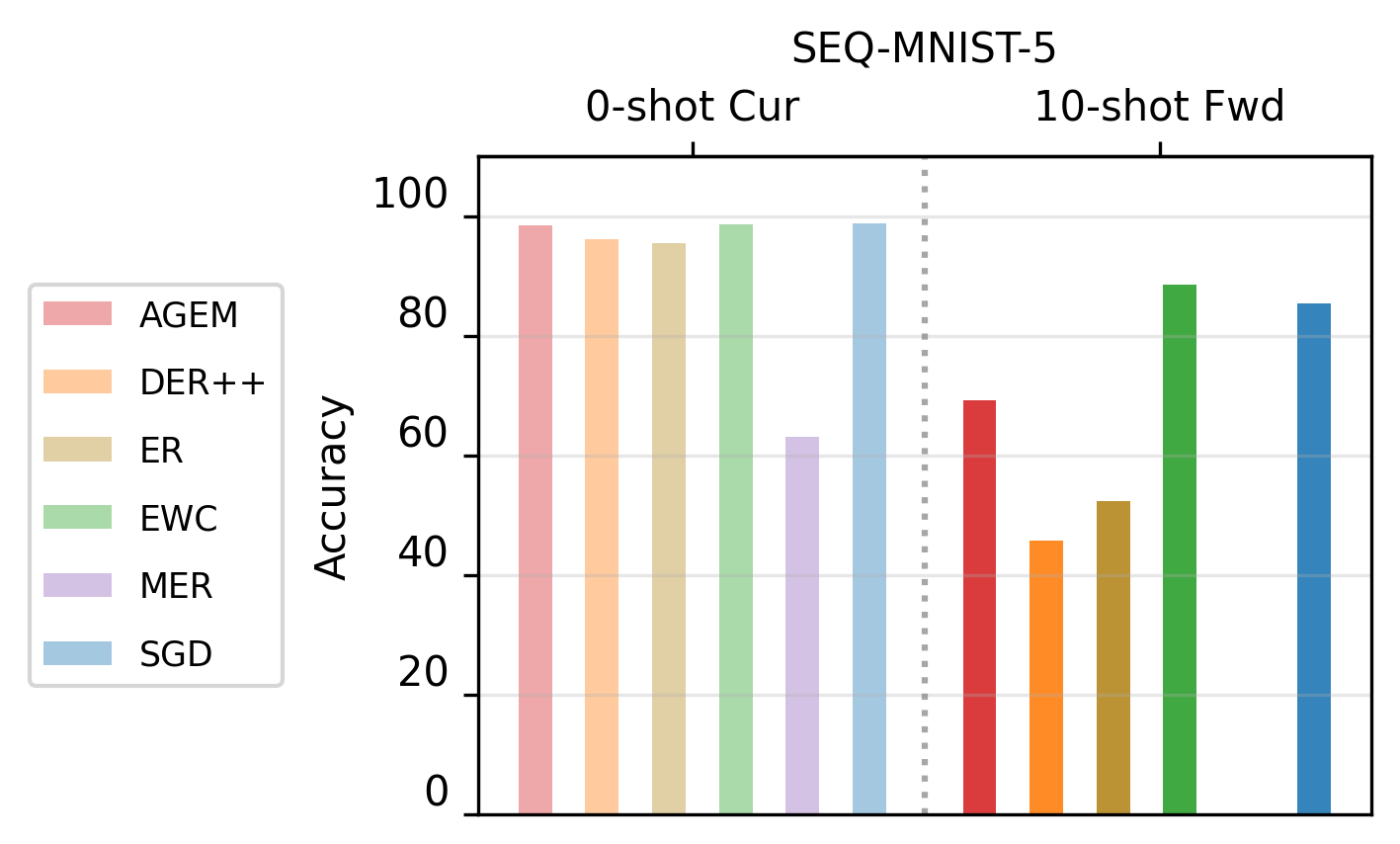}
    \label{fig:plasticity-mnist}
\end{subfigure}
\begin{subfigure}{0.21\textwidth}
        \includegraphics[height=2.8cm]{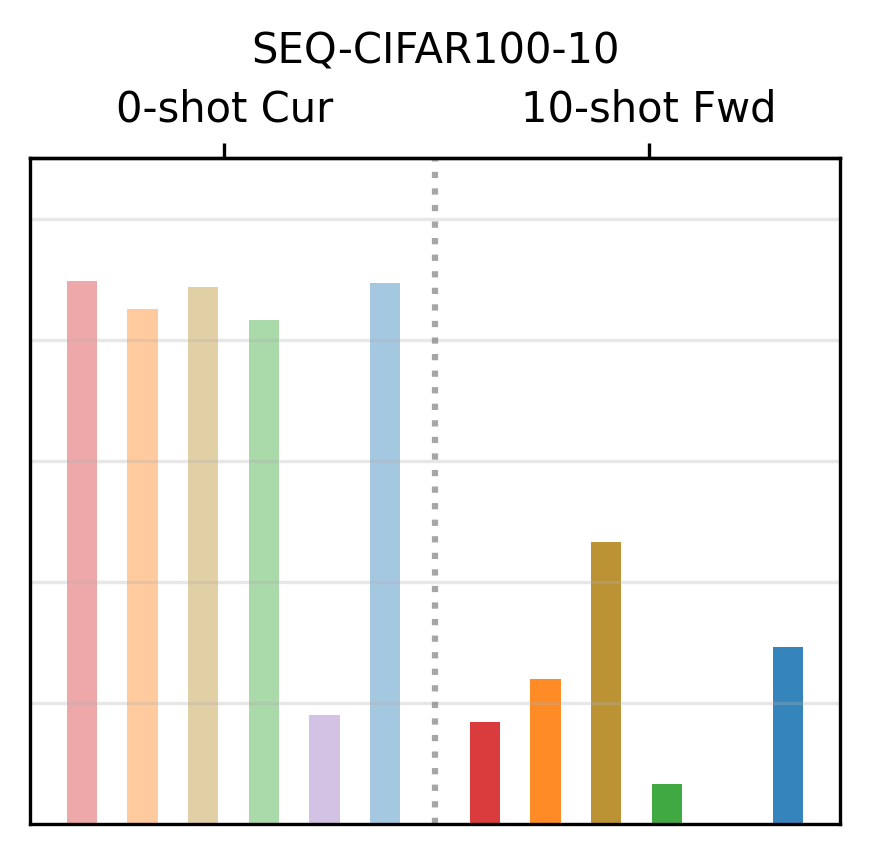}
    \label{fig:plasticity-seq-cifar100}
\end{subfigure}
\begin{subfigure}{0.21\textwidth}
        \includegraphics[height=2.8cm]{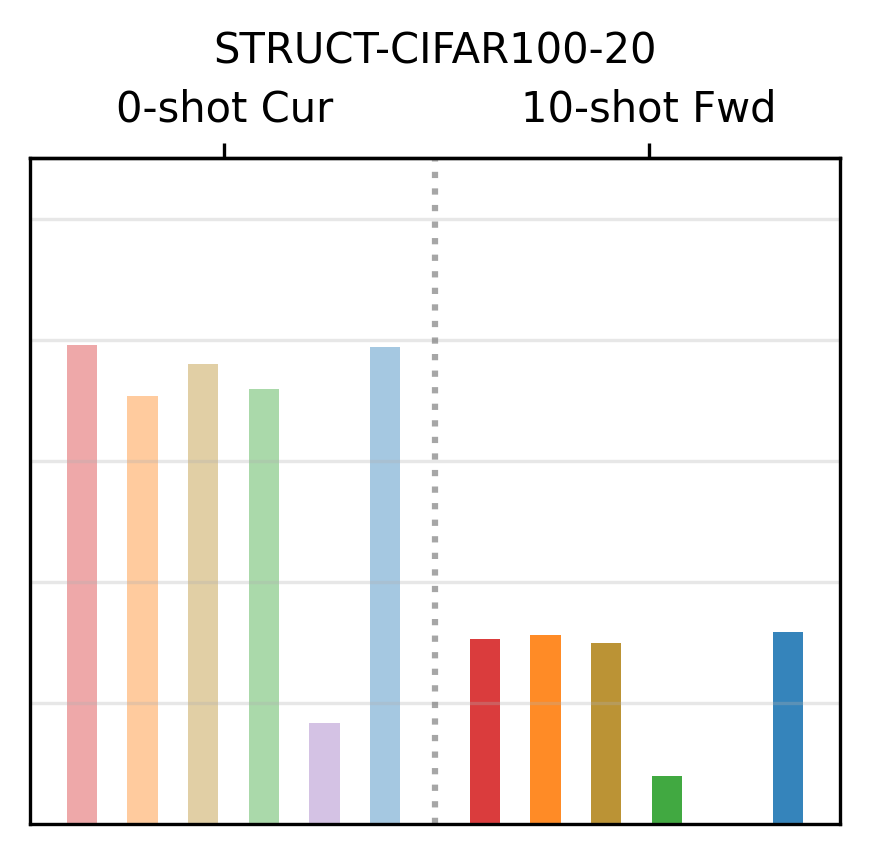}
    \label{fig:plasticity-struct-cifar100}
\end{subfigure}
\begin{subfigure}{0.21\textwidth}
        \includegraphics[height=2.8cm]{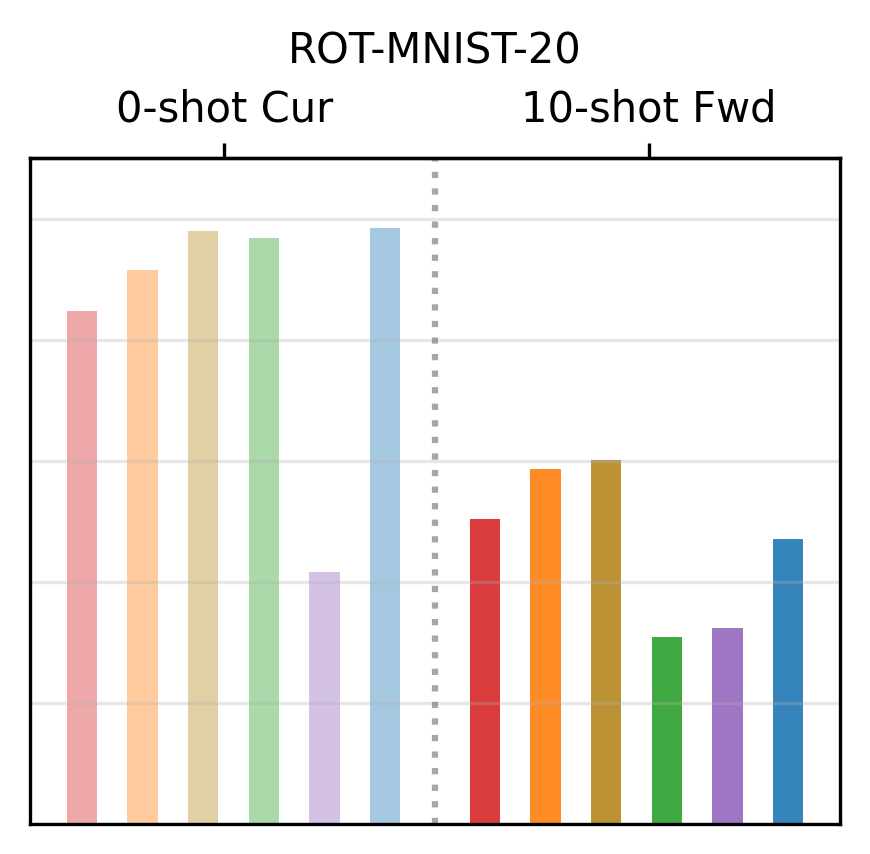}
    \label{fig:plasticity-rot-mnist}
\end{subfigure}
\caption{CL Plasticity (0-shot accuracy on the \emph{current} task) is not highly informative of method performance, as many methods saturate on the most common benchmarks for visual CL. In contrast, few-shot \emph{forward} transfer effectively characterizes model plasticity in visual CL. Evaluated 10-shot, EWC appears to sharply lose plasticity on long sequences, AGEM's plasticity benefits from a lack of cross-task structure, and SGD is highly plastic due to its limited inductive bias.}
\label{fig:plastic-res-1}
\end{figure}

\begin{figure}
\centering
\begin{subfigure}{0.48\textwidth}
        \includegraphics[width=\textwidth]{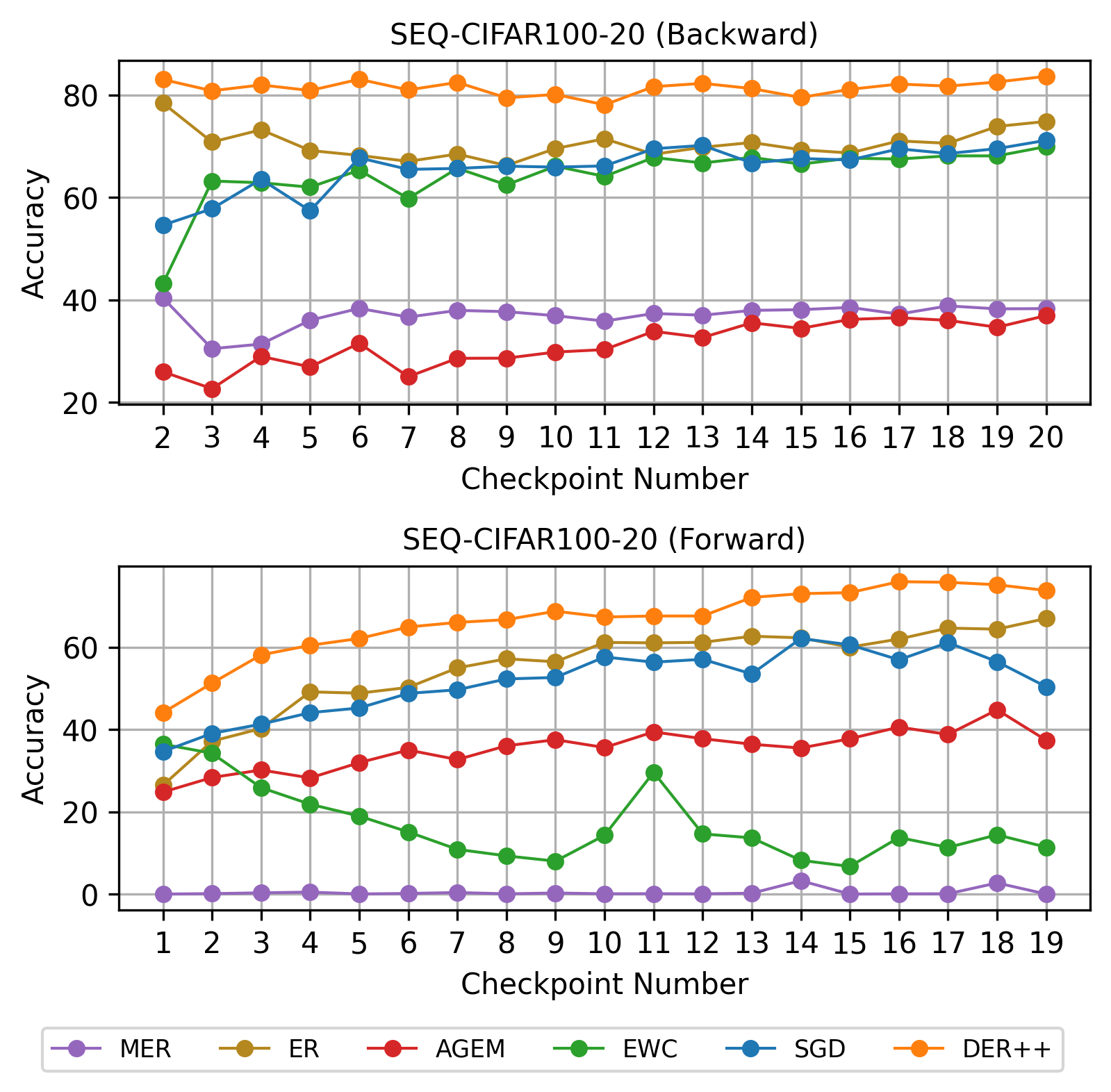}
    \label{fig:improvement-seq-cifar100}
\end{subfigure}
%
\begin{subfigure}{0.48\textwidth}
        \includegraphics[width=\textwidth]{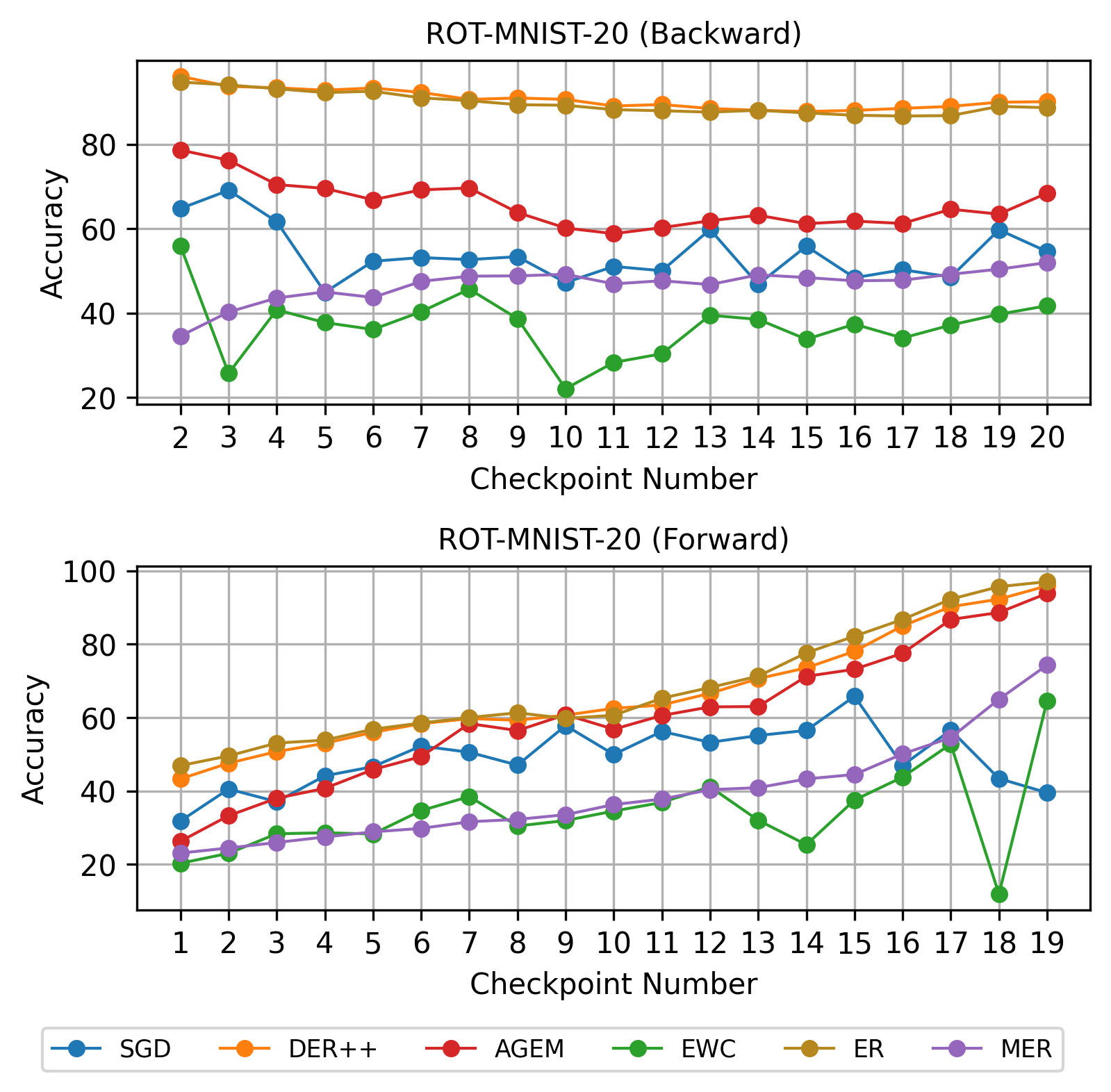}
    \label{fig:improvement-rot-mnist}
\end{subfigure}
\caption{Per-checkpoint backward and forward performance on task- and domain-incremental sequences, after 10-shot adaptation. On \emph{backward} tasks (\textbf{top row}), each method appears to maintain a relatively constant level of stability without deterioration over the task sequence. On \emph{forward} tasks (\textbf{top row}), EWC shows a sharp degradation over time, while SGD and replay-based methods show slight improvements, likely due to their lack of regularization. On {\rotm}, most methods benefit from the high task similarity at the end of the sequence.}
\label{fig:improvement-res-1}
\vspace{-0.5cm}
\end{figure}

Next, we contrast the results with the conventional measure of plasticity (averaged 0-shot current-task accuracy) $\phi^{CL}$ with the 10-shot accuracy on \emph{forward} tasks. In every setting, we observe that $\phi^{CL}$ is roughly identical across methods, leading to the assumption that most methods lead to roughly identical plasticity (Figure \ref{fig:plastic-res-1}, left of each sub-figure). However, examining the 10-shot forward transfer shows that CL methods differ significantly in plasticity as expected from their inductive biases. EWC, which promotes stability via weight-space regularization, shows strong plasticity on the short {\seqm}, but significantly loses plasticity on longer sequences, even with high cross-task structure ({\rotm}). AGEM makes task updates in the null space of prior tasks, and thus shows high plasticity in sequences with low task overlap ({\structc}) and low plasticity otherwise ({\seqc}), even with the same base dataset (CIFAR100). The replay-based methods perform worse with lower task overlap, and MER does not scale well beyond MNIST. Finally, we observe that SGD remains highly competitive with other methods, which is expected as it does not explicitly preserve stability.

Zooming in on these trends, we observe in Figure \ref{fig:improvement-res-1} that the forward performance of AGEM is relatively stable on {\seqct}, and that of SGD and replay-based methods (ER, DER++) improves slightly. Notably, EWC shows a \emph{negative} trend in forward transfer beyond the first few tasks, corroborating our earlier observation of plasticity loss on longer sequences. In our formulation (rotating by $20^\circ$ per task), the first and last two tasks on {\rotm} are similar, leading to positive transfer towards the end of the sequence with every method except EWC.
\section{Improving Plasticity with Foresight Meta-Learning}\label{sec:foresight-meta}

\begin{figure}[h!]
\centering
\begin{subfigure}{0.33\textwidth}
        \includegraphics[height=2.8cm]{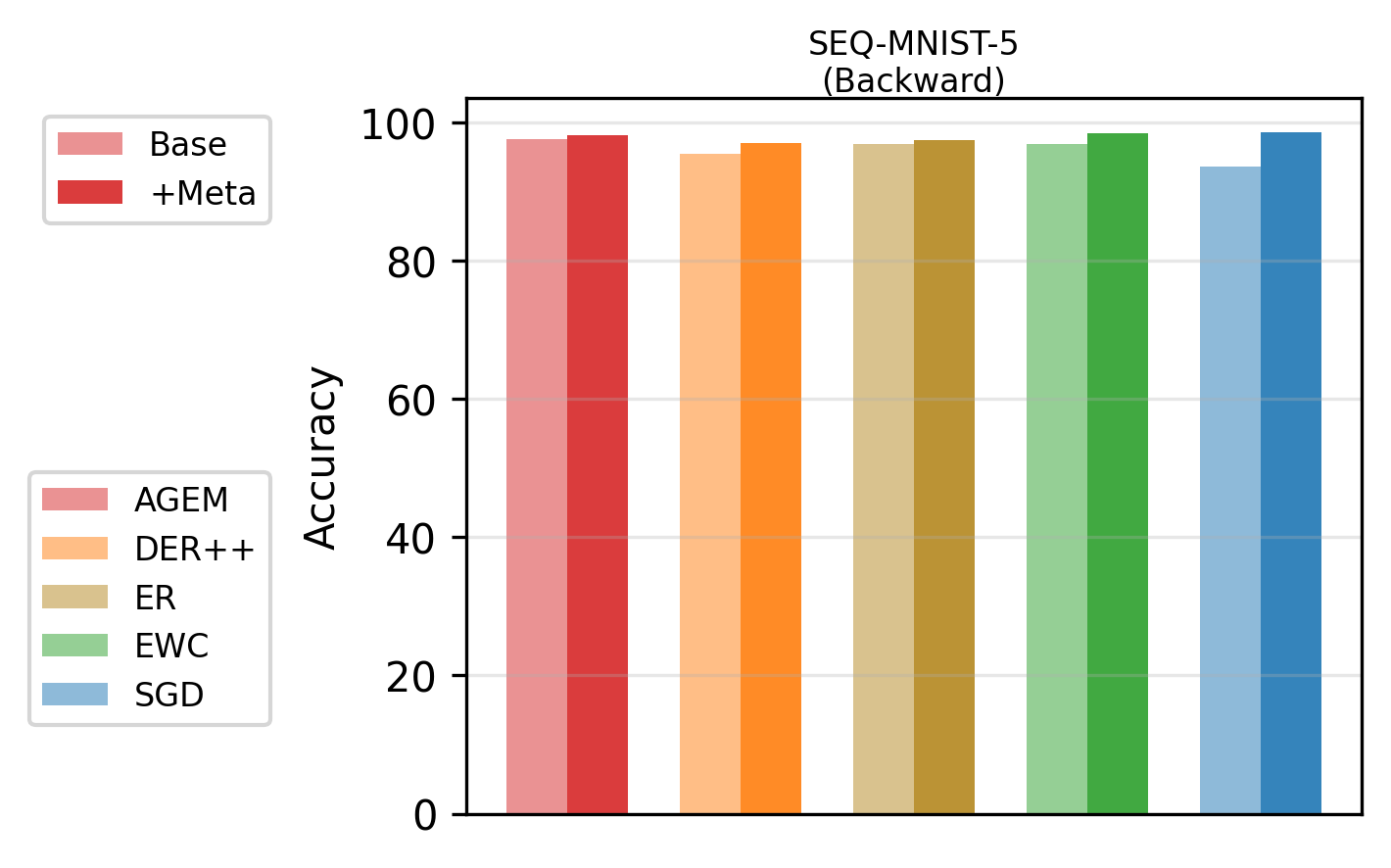}
    \label{fig:meta-stability-mnist}
\end{subfigure}
\begin{subfigure}{0.21\textwidth}
        \includegraphics[height=2.8cm]{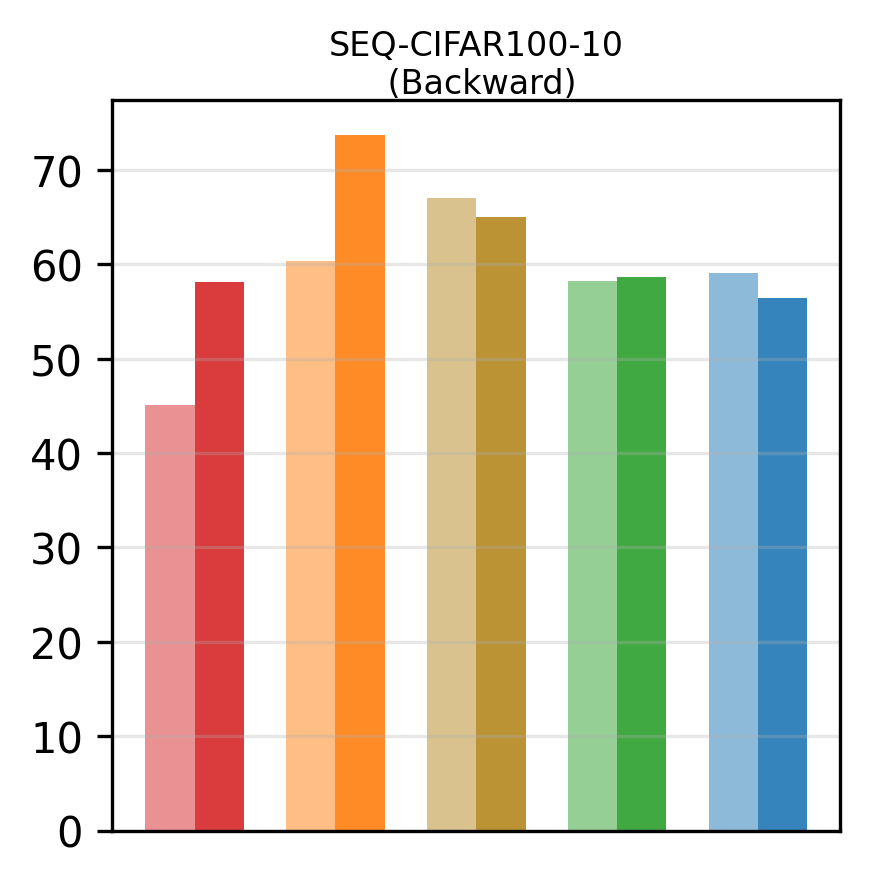}
    \label{fig:meta-stability-seq-cifar100}
\end{subfigure}
\begin{subfigure}{0.21\textwidth}
        \includegraphics[height=2.8cm]{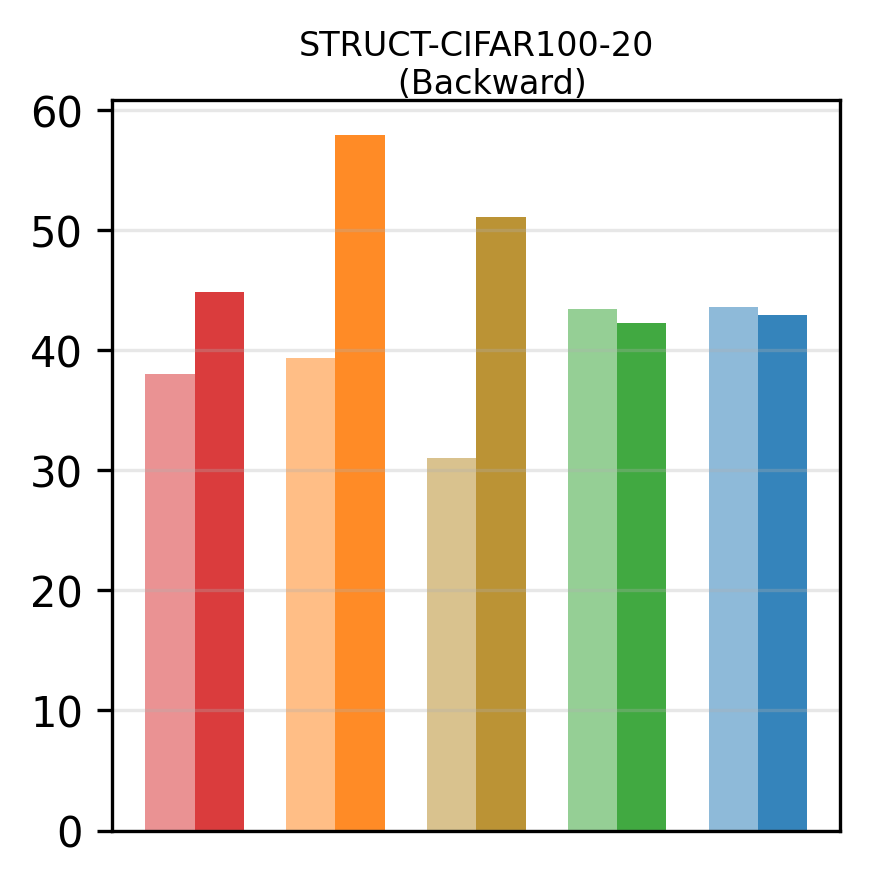}
    \label{fig:meta-stability-struct-cifar100}
\end{subfigure}
\begin{subfigure}{0.21\textwidth}
        \includegraphics[height=2.8cm]{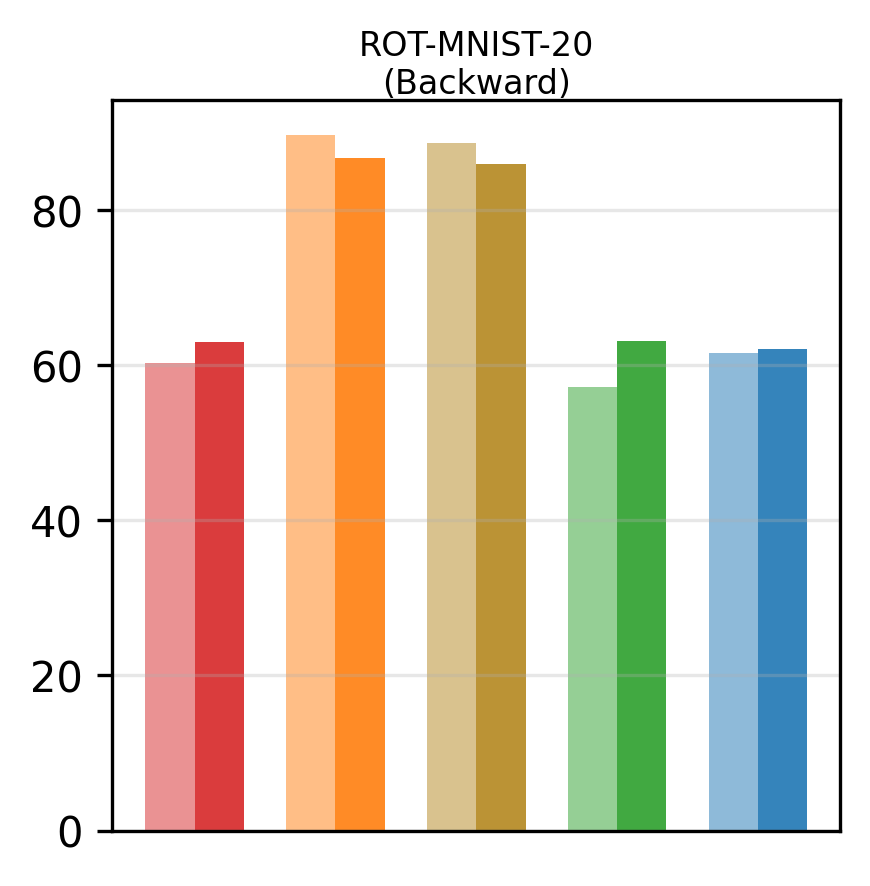}
    \label{fig:meta-stability-rot-mnist}
\end{subfigure}
%
%
\begin{subfigure}{0.33\textwidth}
        \includegraphics[height=2.8cm]{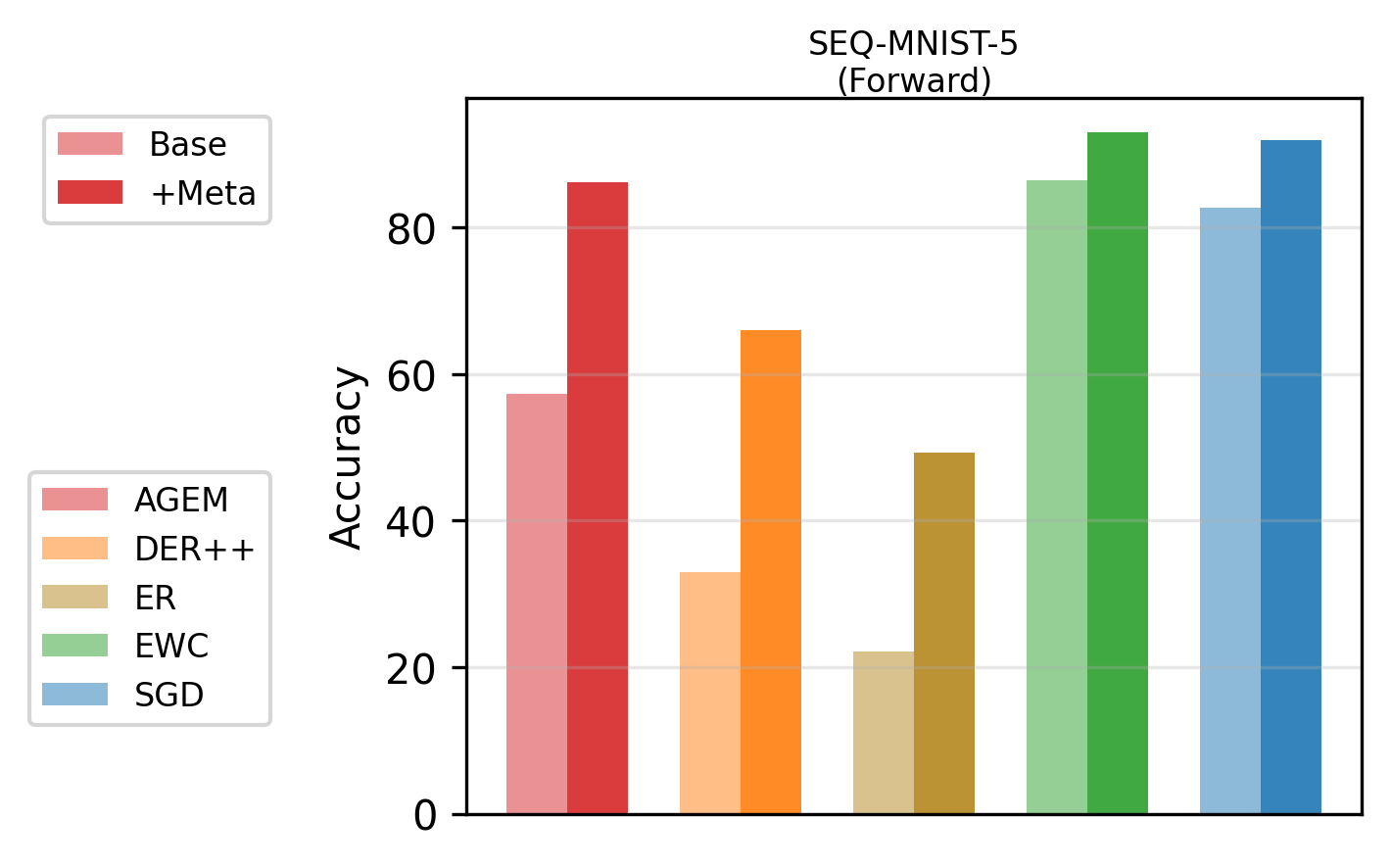}
    \label{fig:meta-plasticity-mnist}
\end{subfigure}
\begin{subfigure}{0.21\textwidth}
        \includegraphics[height=2.8cm]{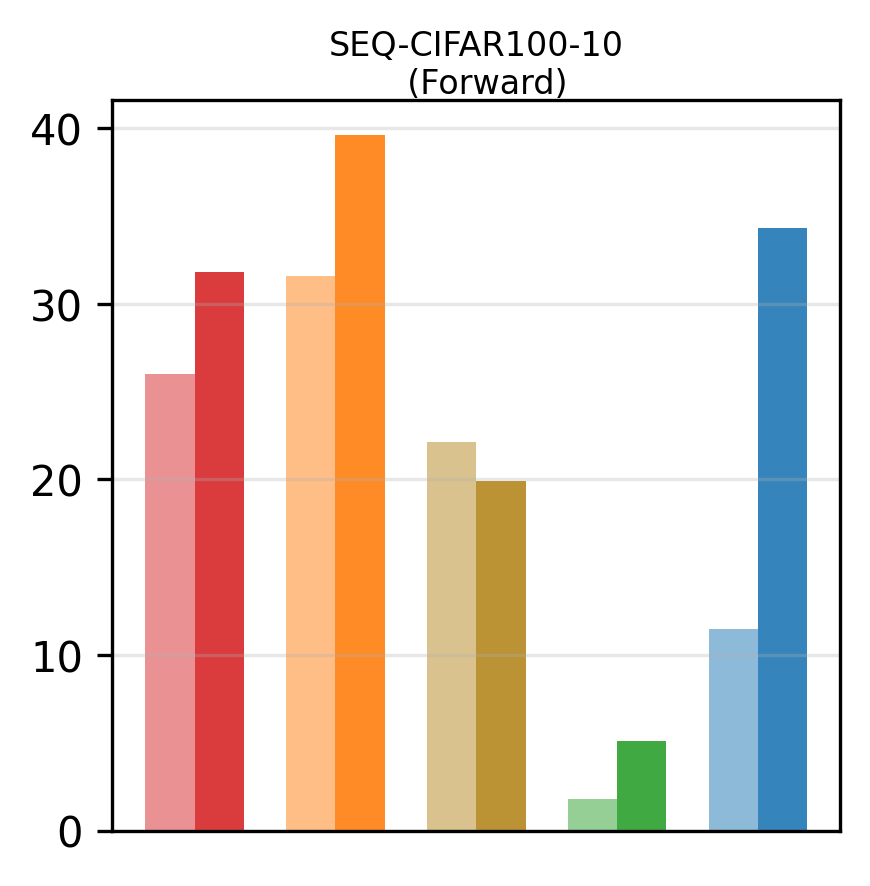}
    \label{fig:meta-plasticity-seq-cifar100}
\end{subfigure}
\begin{subfigure}{0.21\textwidth}
        \includegraphics[height=2.8cm]{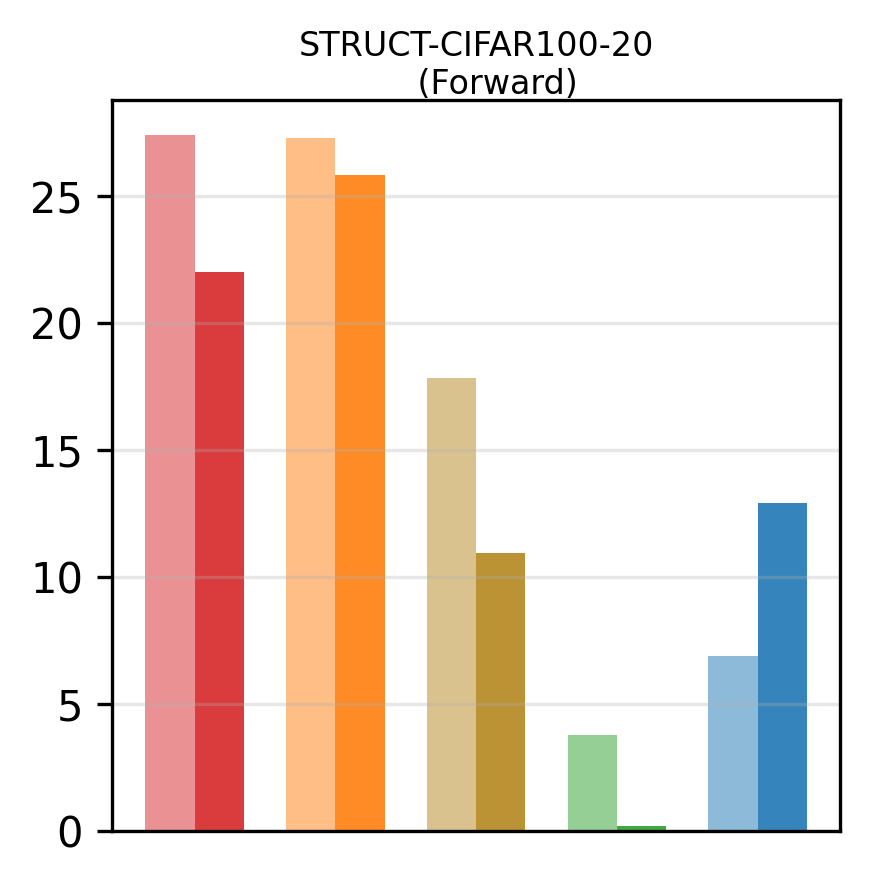}
    \label{fig:meta-plasticity-struct-cifar100}
\end{subfigure}
\begin{subfigure}{0.21\textwidth}
        \includegraphics[height=2.8cm]{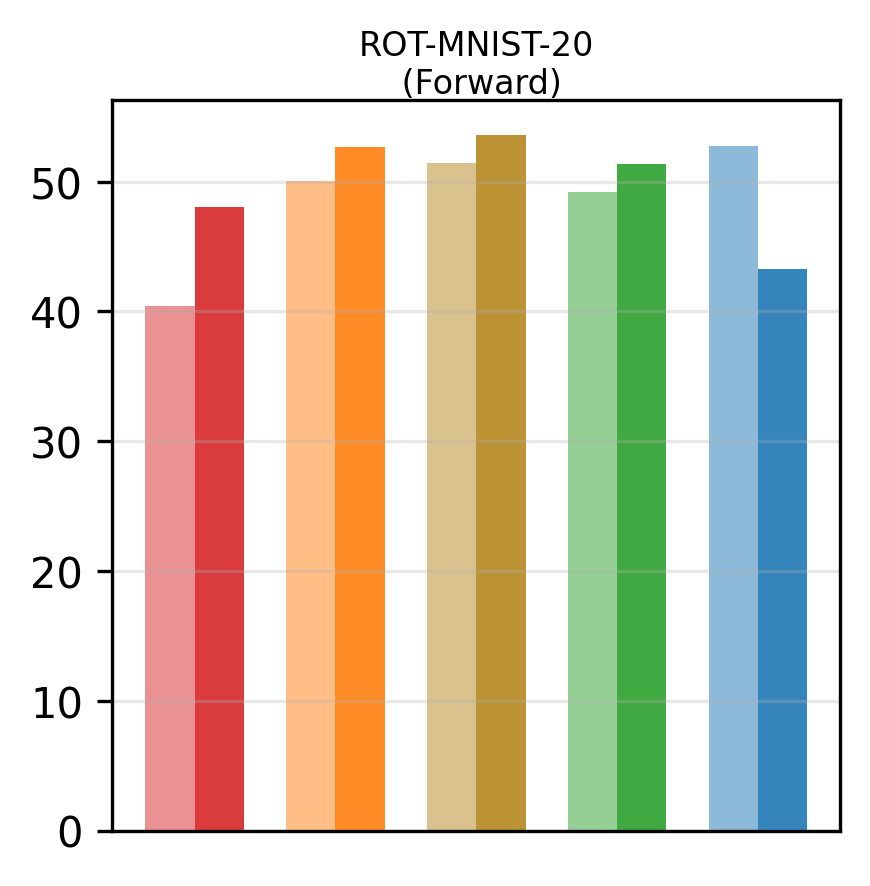}
    \label{fig:meta-plasticity-rot-mnist}
\end{subfigure}
\caption{Performance of 5-shot \emph{backward} (upper row) and \emph{forward} (lower row) transfer with \emph{foresight meta-learning} added to CL methods. We update each method with MAML on 1 ({\seqm}) or 3 (others) look-ahead tasks in parallel (forward measurement \emph{does not} include look-ahead tasks). Many methods see an improvement in forward transfer. Though we only apply meta-learning in the forward direction, we also observe improved backward transfer in some cases.}
\label{fig:meta-cl-res-1}
\end{figure}

In prior sections, we observe that few-shot adaptation can sharply reduce forgetting and characterizes a model's plasticity over the learning sequence, but also that this ability stagnates or worsens over time in existing CL methods. Our results raise a natural question: \emph{can the learning-to-learn ability of a CL model, measured few-shot, be optimized as the model progresses through the learning sequence?}

As we observe in Section \ref{related:meta-learn}, existing methods for meta-CL optimize for stability and not rapid adaptation, as their meta-updates are \emph{backward-looking}. Ideally, an algorithm should learn to adapt in the \emph{forward} direction to perform well on a continuously evolving sequence. As a proof of concept, we develop a method in a \emph{rolled back} CL setting, where at task $t+L$, we train a learner that has learned only up to task $t$ (i.e., $\theta_t$ are `slow weights', Figure \ref{fig:intro-fig}). Instead of a replay buffer, we leverage up to $k$ \emph{unseen} examples from tasks $\mathcal{T}_{t+1},\ldots,\mathcal{T}_{t+L}$ to meta-learn the task sequence prior to training on the full task $\mathcal{T}_{t+1}$. To distinguish this method from existing works that meta-learn from examples that were already learned, which can be considered hindsight meta learning,  we refer to our method as adding \emph{foresight meta-learning} to a CL method. The standard CL `outer loop' trains a model many-shot on a task sequence:
\begin{align}
    \theta_S &= U_N^{S}\circ U_N^{S-1}\circ\cdots\circ U_N^1(\theta_0)
\end{align}
Here, $U_N^{j}(\theta)=U_N(\theta,\mathcal{T}_j^\mathrm{train})$ is the many-shot optimization procedure of training $\theta$ on at most $N$ samples of training data from $\mathcal{T}_i$ with some CL strategy $\Omega$. To this procedure, we add an `inner loop' that uses a meta-learning algorithm such as MAML \cite{finn2017modelagnosticmetalearningfastadaptation} or Reptile \cite{nichol2018firstordermetalearningalgorithms} to initialize the model each look-ahead task
At task $t$ in the sequence, our optimization with MAML is the following:
\begin{align}
    \theta_{t}' &= \theta_{t} - \beta\nabla_\theta\sum_{i=t+1}^{\min(t+L,S)} \mathcal{L}_{i}(\mathcal{T}_i^\mathrm{val}, U_k(\theta_{t}, \mathcal{T}_i^\mathrm{train})) \\
    \theta_{t+1} &= U_N(\mathcal{T}_{t+1}^\mathrm{train}, \theta_{t}') 
    \label{eqn:meta-parallel}
\end{align}
The inner loop meta-learns an initialization $\theta_{t}'$ that optimizes for few-shot performance on future tasks, and the outer loop trains model on task $t+1$ to obtain $\theta_{t+1}$. 
We also propose a variant of foresight meta-learning that leverages the structure of the CL paradigm. Within the inner loop, each look-ahead model is trained on a \emph{sequence} of future tasks $\mathbb{S}$.
\begin{equation}
    \begin{aligned}
        \mathbb{S} &= \Phi(\{\mathcal{T}_{t+1},\ldots,\mathcal{T}_{t+L}\})\\
        \theta_{t}' &= \theta_{t} - \beta\nabla_\theta\sum_{S_i\in \mathbb{S}} \mathcal{L}_{i}(\mathcal{S}_i^\mathrm{val}, U_k^{S_i^L}\circ\cdots\circ U_k^{S_i^1}(\theta)) \label{eqn:many-seq}
    \end{aligned}
\end{equation}
In our experiments, $\mathbb{S}$ is a singleton set corresponding to the CL sub-sequence $\mathcal{T}_{t+1}\rightarrow\cdots\rightarrow\mathcal{T}_{t+L}$. In scenarios where the task ordering is not crucial to learning, it may be beneficial to add more sequences to $\mathbb{S}$ (via, e.g., permutation) for a better meta-learning signal.

\begin{figure}[h!]
\centering
%
%
%
\begin{subfigure}{0.8\textwidth}
        \includegraphics[width=\textwidth]{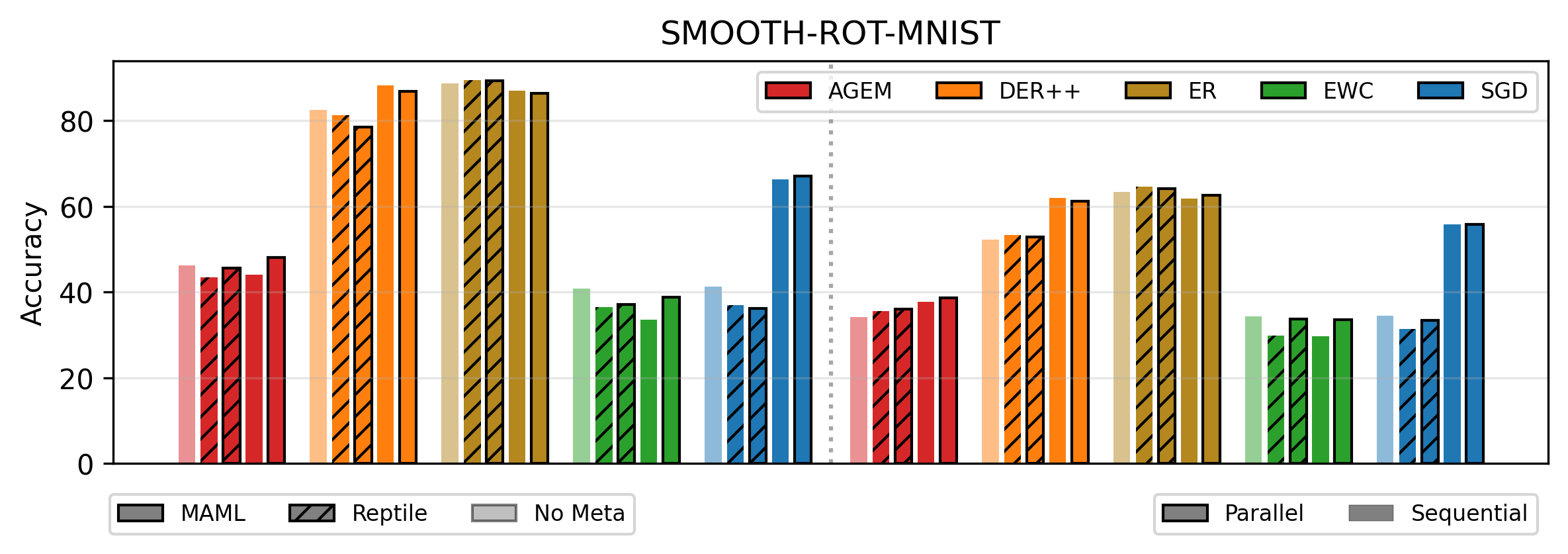}
    \label{fig:meta-methods-rot-mnist}
\end{subfigure}
\caption{5-shot backwards and forwards accuracy of \emph{foresight meta-learning} methods on {\rotm}, varying on the meta-learning algorithm: (MAML/Reptile/fine-tuning) and the update strategy: (parallel/sequential). Adding MAML outperforms fine-tuning on look-ahead examples.}
\label{fig:meta-cl-res-2}
\vspace{-0.5cm}
\end{figure}

Figure \ref{fig:meta-cl-res-1} shows that adding \emph{foresight meta-learning} improves the performance on forward tasks beyond the look-ahead window for most methods on each task sequence. The exception to this is {\structc}, where the lack of shared task structure appears to lead to worse performance with meta-learning.
Notably, this procedure even results in improved \emph{backward} transfer despite no direct optimization for this objective, indicating that the model's general ability to adapt may be improved. We see further evidence for this in the following section, where we study plasticity as the ability to \emph{rapidly adapt to a task}, in alignment with studies on natural learners. We present an example of our method on {\rotm} (Figure \ref{fig:smooth-mnist-maml-parallel-full}), as well as full results on all methods (Figure \ref{fig:meta-cl-res-full}) and a detailed comparison of meta-SGD variants with error bars (Figure \ref{fig:smooth-mnist-meta-sgd-full}) in Appendix \ref{apx:more-results}.

We evaluate the differences between meta-learning strategies on {\rotm}, a sequence with high task similarity. Figure \ref{fig:meta-cl-res-2} shows a comparison between the parallel (Eqn \ref{eqn:meta-parallel}) and sequential (Eqn \ref{eqn:many-seq}) versions of \emph{foresight meta-learning} with either MAML \cite{finn2017modelagnosticmetalearningfastadaptation}, Reptile \cite{nichol2018firstordermetalearningalgorithms}, or a baseline that simply fine-tunes on look-ahead examples. As expected, the second-order MAML generally outperforms other strategies, though Reptile performs well with ER. On EWC (the weakest forward-transfer method), we observe that meta-learning adds limited benefit. The sequential strategy performs similarly to parallel, and we hypothesize that differences require the study of longer task sequences.

\section{Measuring the Rate of Adaptation with Per-shot Plasticity}\label{sec:sauce}

\begin{figure}[ht]
\centering
\centerline{\includegraphics[width=0.9\linewidth]{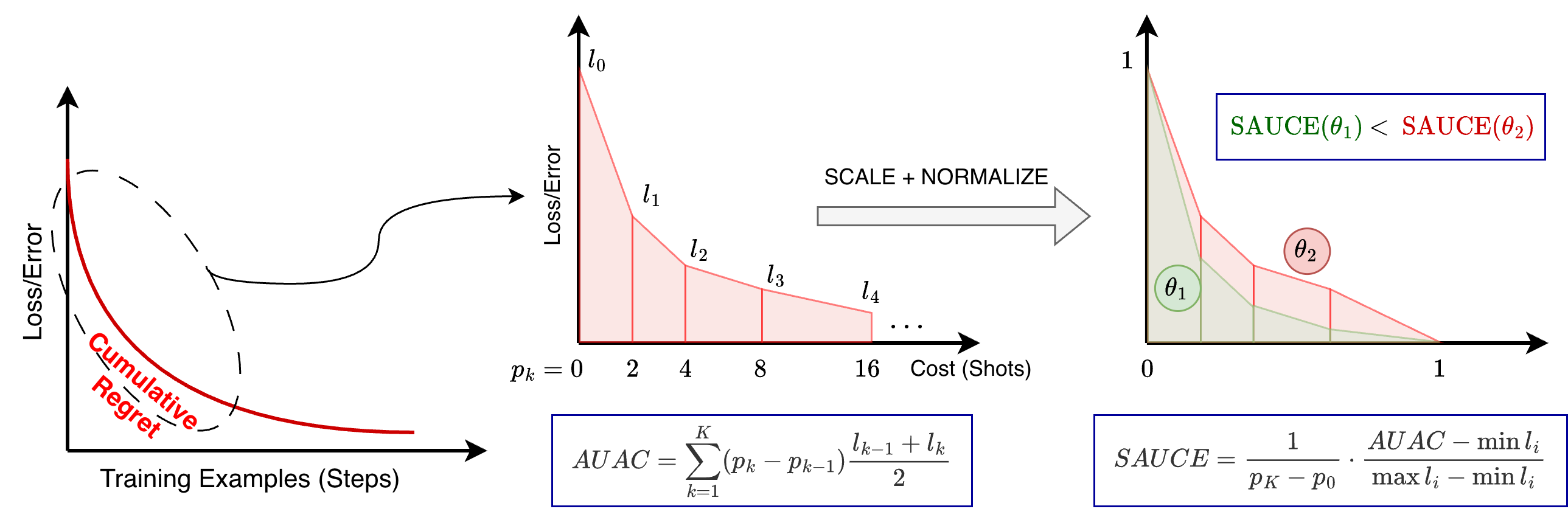}}
\caption{
  A metric that captures the rate of adaptation as the scaled regret over early training.
  \textbf{Left--Middle}: The error after $k$-shot adaptation at each value of $k$ represents a point on the training loss curve. The AUAC measures the total error up to $k=K$ (i.e., `regret'). \textbf{Right}: SAUCE measures the relative improvement with each shot. $\theta_1$ adapts quicker than $\theta_2$ as $\text{SAUCE}(\theta_1)<\text{SAUCE}(\theta_2)$.
}
\label{fig:AUAC}
\vspace{-0.5cm}
\end{figure}

A major consideration in the few-shot evaluation paradigm is the selection of an `appropriate' value of $k$ for $k$-shot adaptation. This choice is driven by multiple factors: the tasks, model, and methods being evaluated; the amount of adaptation required to differentiate between methods; and the number of examples that can practically be obtained and learned from in deployment. A potential solution to this issue is to aggregate information across multiple values of $k$ (effectively, the LCA \citep{AGEM}), but this fails to capture an important aspect of plasticity---the \emph{rapidity} of adaptation to increasing information. 

Consider two models $\theta_1,\theta_2$ that are evaluated on a task $\mathcal{T}$ after adaptation with $k=0,\ldots,K$ task examples.
Even with identical 0- and $K$-shot performance, $\theta_1$ may be much quicker to adapt to the task than $\theta_2$, rapidly improving with only 1 or 2 examples. This would make $\theta_1$ the preferred choice in scenarios with rapidly changing distributions or limited training resources. We define a novel metric that captures this adaptability: the \emph{per-shot plasticity}. Let $A=\{a_1,\ldots,a_K\}$ be a set of few-shot adaptation strategies that can be applied to a model $\theta$ to improve performance. We define $l_j=l_\mathcal{T}(\theta, a_j)=\mathbb{E}_{(x,y)\sim\mathcal{T}}[\mathcal{L}(f_\theta(a_k||x),y)]$ to be the expected loss of $\theta$ adapted with $a_k$ on task $\mathcal{T}$. We quantify the cumulative error over the set $A$ as \emph{Area Under the Performance-Adaptation Curve (AUAC)} (Figure \ref{fig:AUAC}, middle). This curve measures the absolute decrease in task error with increasing adaptation. For adaptation with gradient descent (as in our setting), this approximates the curve of training error with a batch size of 1, which represents the \emph{early regret} \cite{blackwell1956analog} in learning.
\begin{equation}\label{eqn:AUAC}
    \mathrm{AUAC}(\theta, A, p, \mathcal{T}) = \sum_{k=1}^{K}(p_{k+1}-p_k)\cdot \frac{l_k + l_{k+1}}{2}
\end{equation}
Here, $p_k=p(a_k)$ represents a complexity-dependent weighting of increasing adaptation which we set to $k$ (the number of shots) for simplicity. While the AUAC aggregates the absolute error over early learning, its numerical value is highly dependent on the 0- and $K$-shot performance of a model.
To isolate the \emph{rate of adaptation} specifically, we scale the worst performing adaptation to an error of 1, and the best to 0. We also scale by the adaptation range $(p_K-p_0)$ to produce a metric that is numerically in [0,1]. We call this metric the \emph{Scaled Area Under the Adaptation CurvE} or SAUCE.
\begin{equation}\label{eqn:SAUCE}
    \mathrm{SAUCE}(\theta, A, p, \mathcal{T}) = \frac{1}{p_k-p_0}\cdot\frac{\mathrm{AUAC}(\cdots) - \min_{a_j\in A} l_j}{\max_{a_j\in A} l_j - \min_{a_j\in A} l_j}
\end{equation}
SAUCE measures the \emph{relative improvement} in the performance of a model with increasing adaptation. We then define the \emph{backward per-shot plasticity} as the aggregated SAUCE over over backward tasks, akin to \emph{cued recall} in human learners. On future tasks, this metric is the \emph{forward per-shot plasticity}, measuring the \emph{learning-to-learn} ability at a given point in the sequence.

\begin{figure}
\centering
\begin{subfigure}{0.33\textwidth}
        \includegraphics[height=5.8cm]{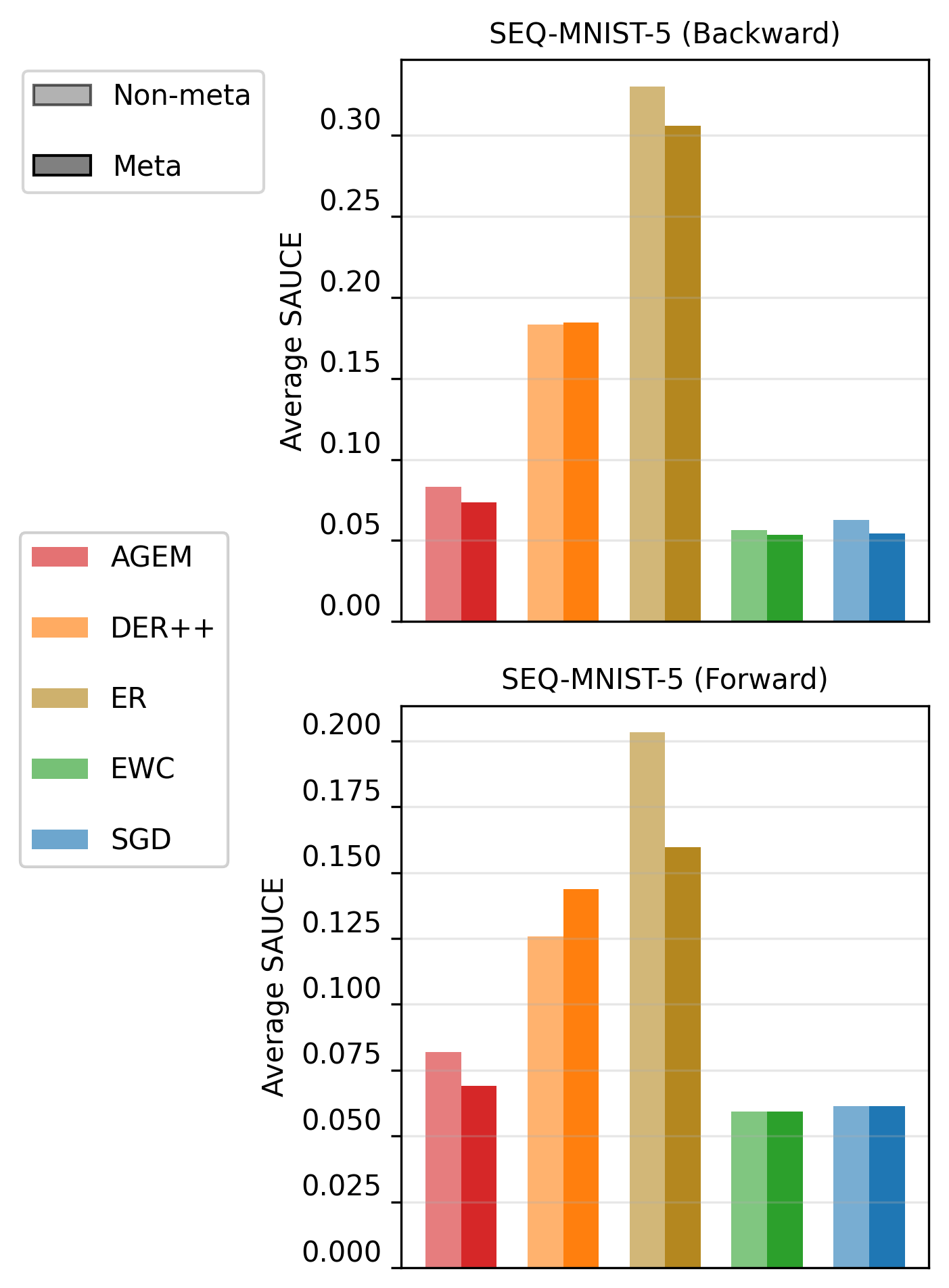}
    \label{fig:sauce-avg-mnist}
\end{subfigure}
\begin{subfigure}{0.21\textwidth}
        \includegraphics[height=5.8cm]{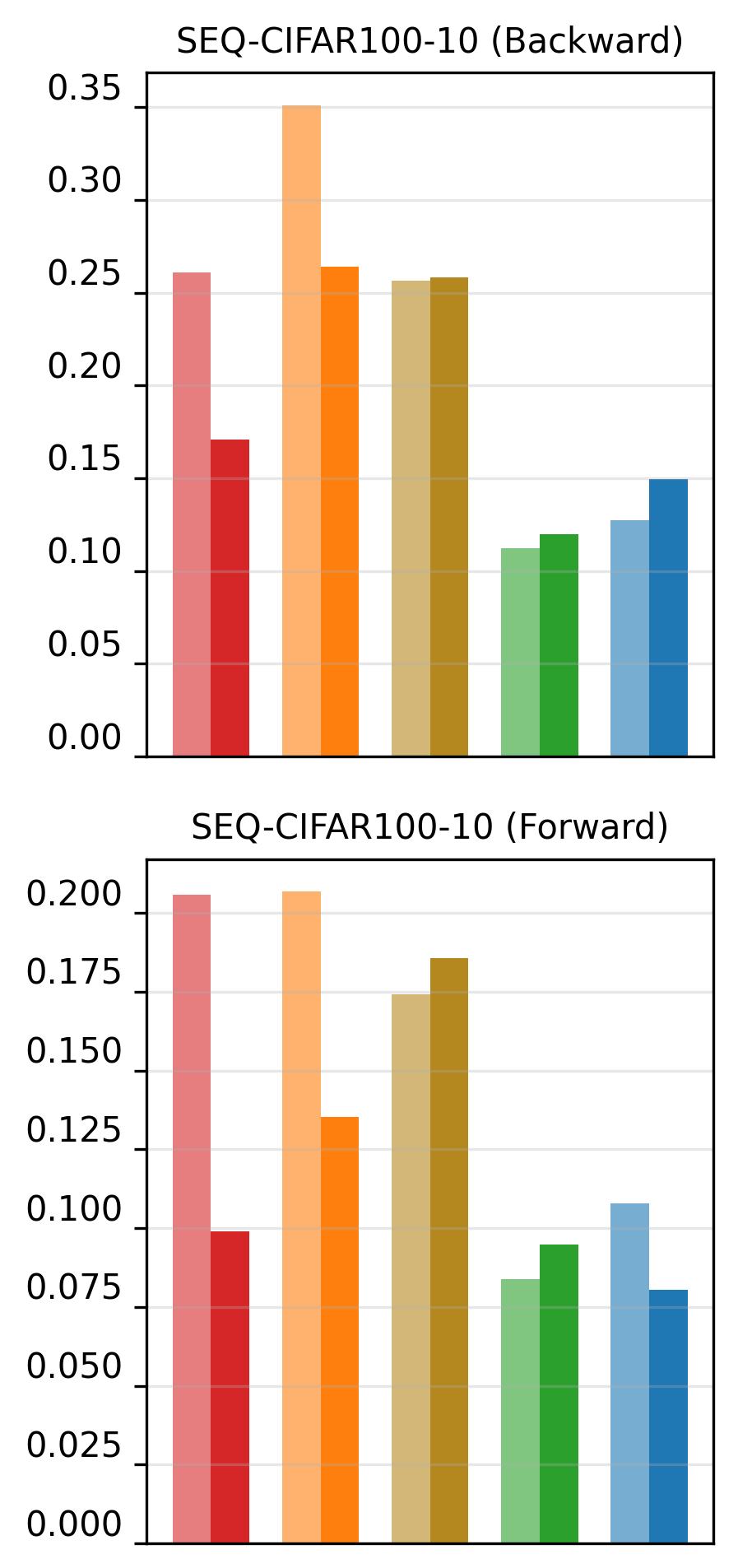}
    \label{fig:sauce-avg-cifar100}
\end{subfigure}
\begin{subfigure}{0.21\textwidth}
        \includegraphics[height=5.8cm]{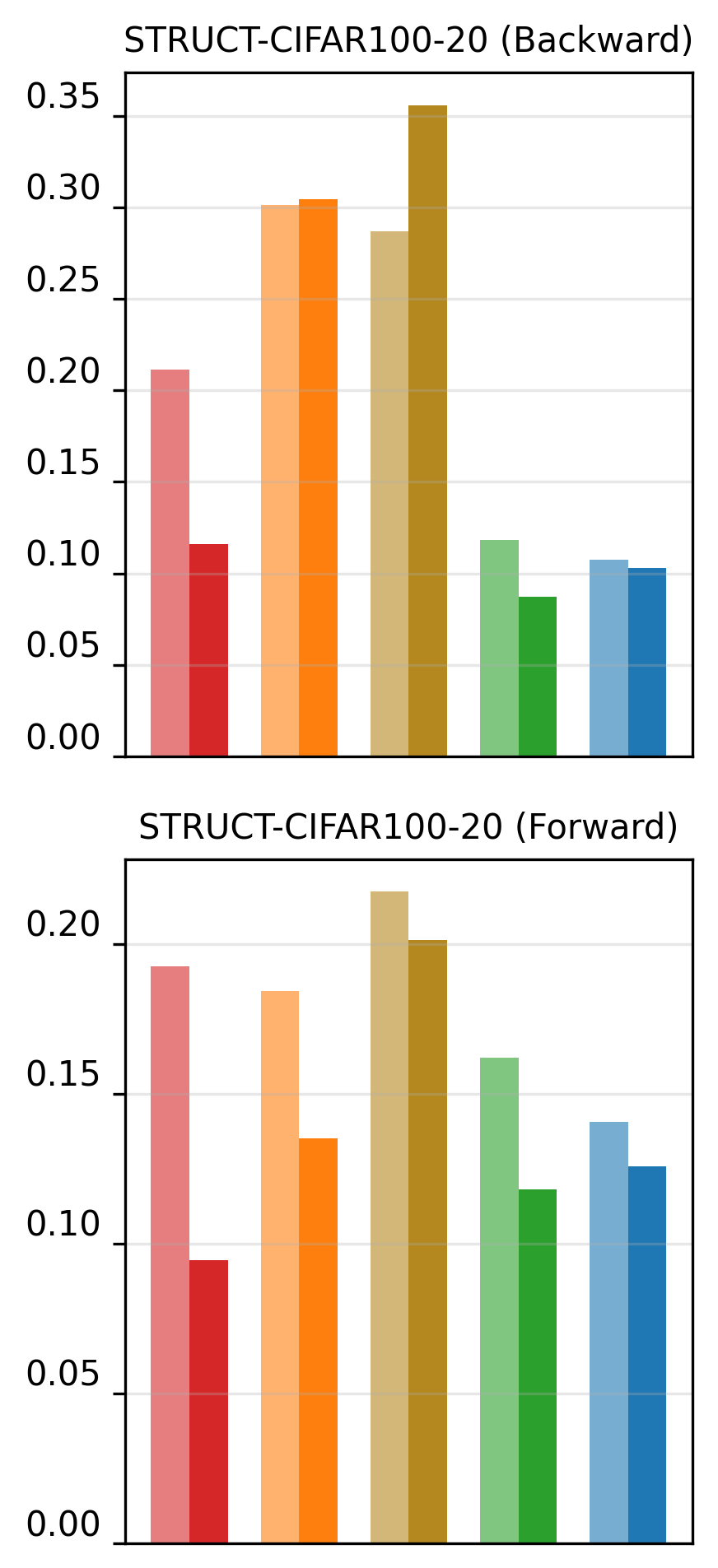}
    \label{fig:sauce-avg-cifar100}
\end{subfigure}
\begin{subfigure}{0.21\textwidth}
        \includegraphics[height=5.8cm]{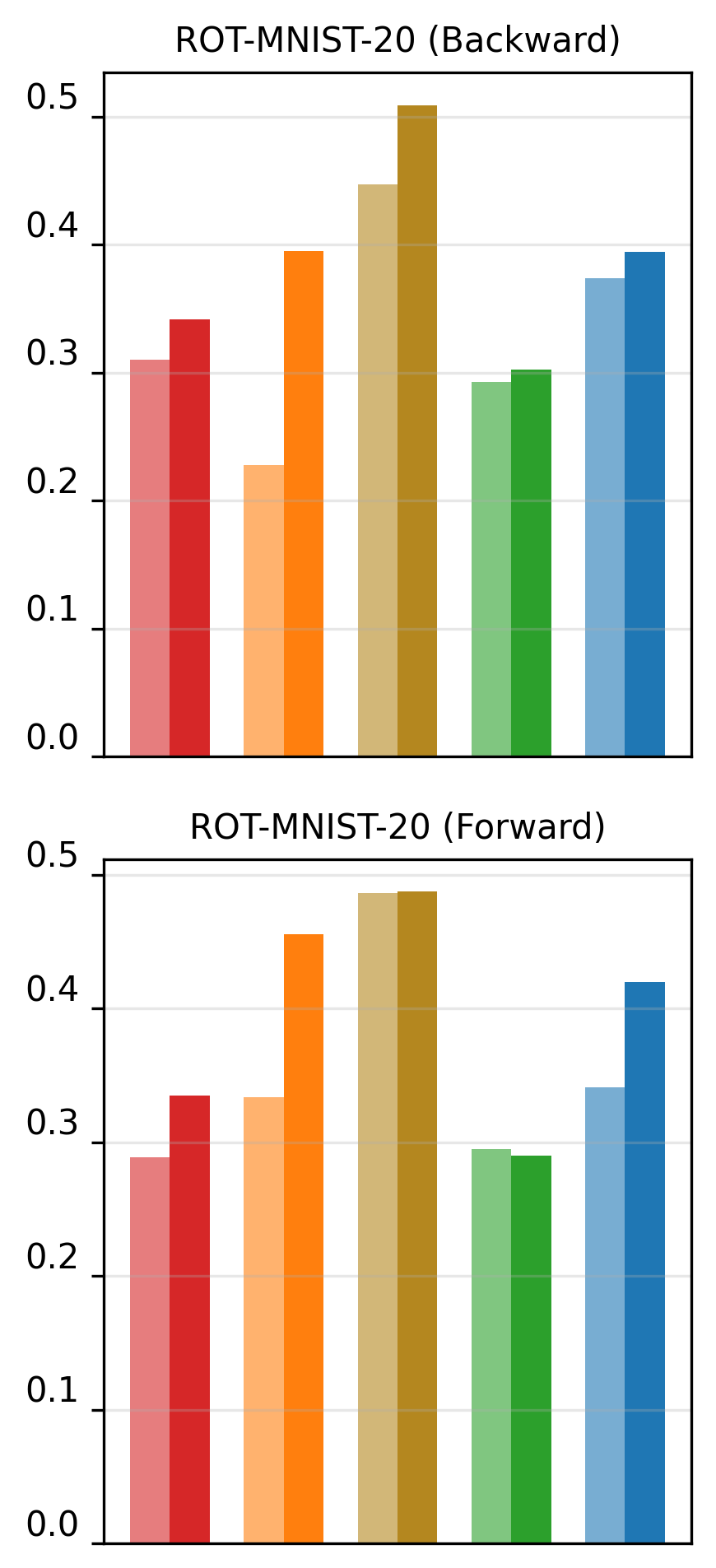}
    \label{fig:sauce-avg-mnist}
\end{subfigure}
\caption{Forward and backward per-shot plasticity, measured by averaged SAUCE, for each method in our setting. +Meta indicates the addition of \emph{foresight meta-learning} to a method. \textbf{Lower} is better. Adding meta-learning improves per-shot plasticity across multiple settings (especially for AGEM), particularly in scenarios with low task similarity (e.g., on {\structc}).}
\label{fig:pershot-res-2}
\end{figure}

Figure \ref{fig:pershot-res-2} shows the average SAUCE over the forward and backward tasks of each checkpoint in our experiment, averaged over checkpoints. As expected, SGD remains the most plastic method due to its lack of additional inductive bias. However, replay-based methods (ER, DER++), which perform very well in the few-shot setting, appear to \emph{reduce} the ability of models to rapidly adapt to new tasks compared to the SGD baseline and regularization-based methods. Thus, in settings with very few examples for adaptation, a replay-based method that has a high performance ceiling but is sample-inefficient may not be the right choice for practitioners. Such an insight does not emerge with conventional metrics for plasticity, highlighting the practical benefit of SAUCE as an evaluation metric.
We find that adding \emph{foresight meta-learning} to most methods leads to an improvement in rapid adaptation ability by a small margin on average. The greatest improvement appears to be for the AGEM method.
However, to truly exhibit \emph{learning-to-learn behavior}, a model's rate of adaptation should improve over the continual learning of the task sequence.

\begin{figure}[ht]
\centering
%
\begin{subfigure}{0.46\textwidth}
        \includegraphics[width=\textwidth]{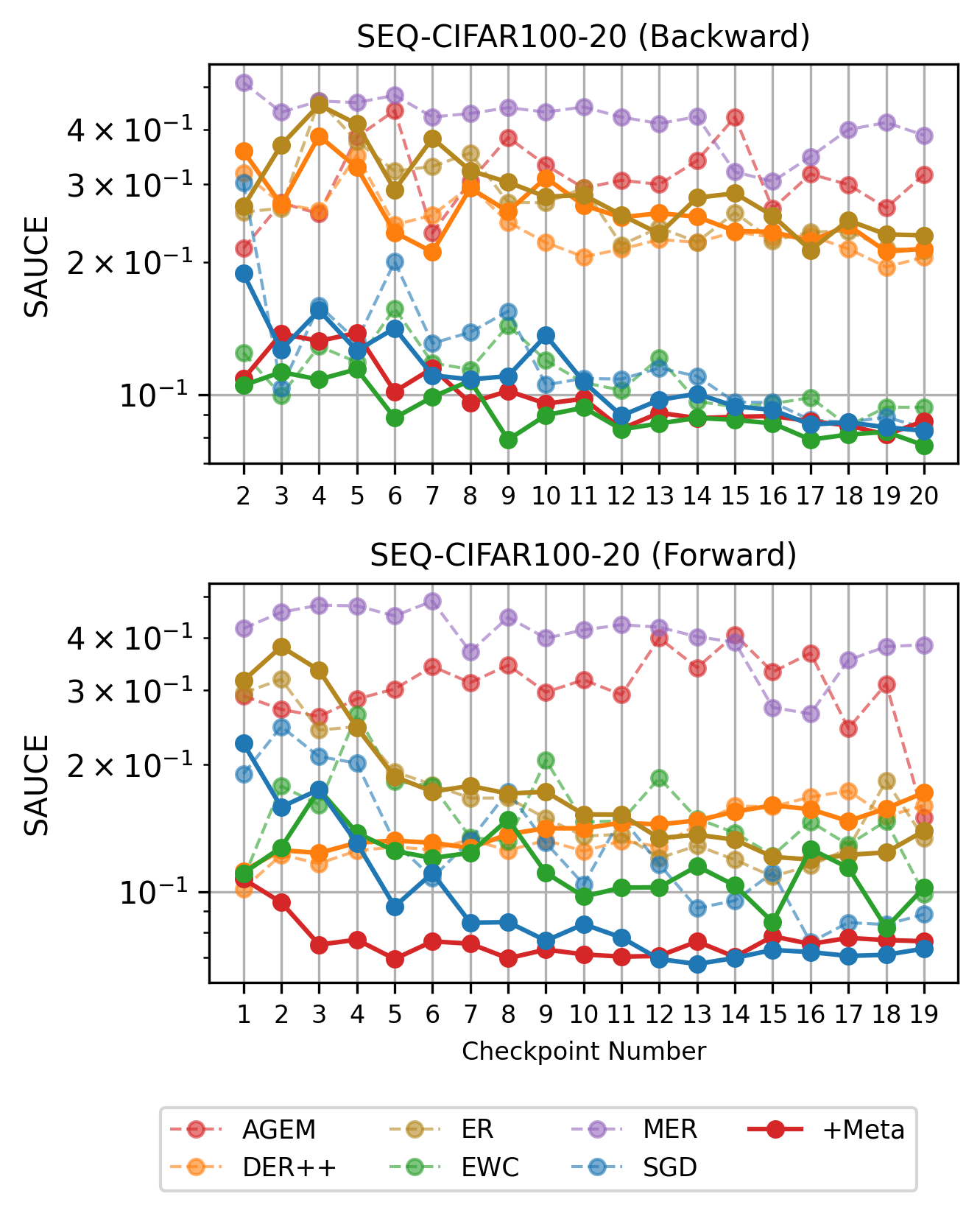}
    \label{fig:sauce-seq-cifar100-20}
\end{subfigure}
\begin{subfigure}{0.46\textwidth}
        \includegraphics[width=\textwidth]{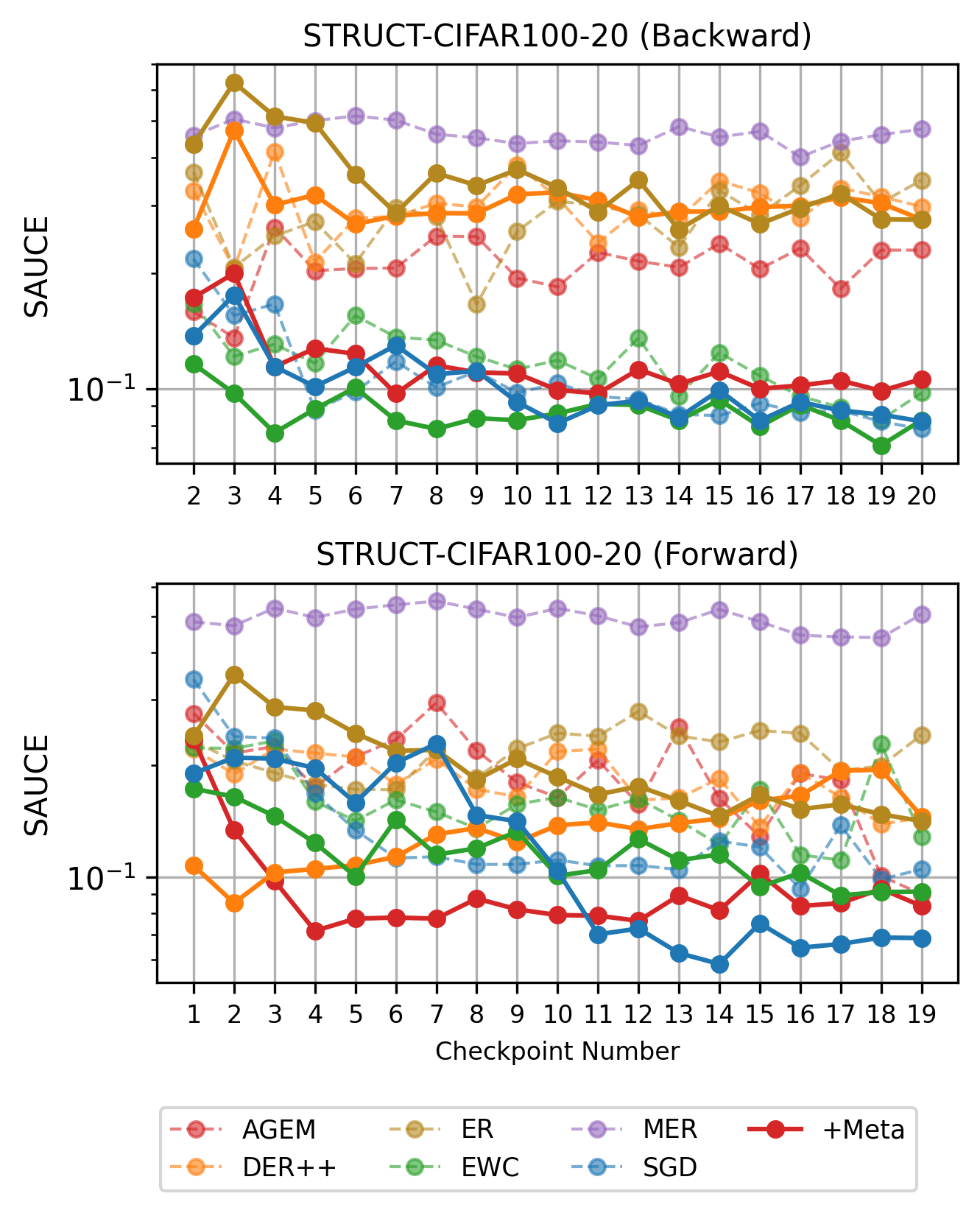}
    \label{fig:sauce-struct-cifar100}
\end{subfigure}
%
\caption{Forward and backward per-shot plasticity on 20-task sequences. The solid lines represent methods augmented with \emph{foresight meta-learning}. The adaptation rate (SAUCE) remains relatively static across existing methods across sequences, but methods augmented with foresight meta-learning show \emph{learning-to-learn} behavior as adaptivity increases over the sequence.}
\label{fig:pershot-res-1}
\end{figure}

Figure \ref{fig:pershot-res-1} shows the average backward and forward per-shot plasticity (measured by SAUCE) for each continual learning method on long task sequences. Between non-meta methods (dashed lines) on {\seqct}, EWC and the SGD baseline show the highest forward per-shot plasticity on the sequences with task overlap, with MER performing the worst despite being motivated by gradient alignment (this is likely due to the small batch size of the method). The rate of adaptation appears to be relatively flat, indicating minimal \emph{learning-to-learn} behavior over the sequences. Without task overlap ({\structc}), AGEM shows the most rapid improvement and best final plasticity, which we attribute to its inductive bias of partitioning  task representations into orthogonal subspaces. On the domain-incremental {\rotm} sequence, all the models perform equivalently, with relatively flat per-shot plasticity across tasks.

On every sequence except for {\rotm}, adding \emph{foresight meta-learning} to a method (solid lines in Figure \ref{fig:pershot-res-1}) not only improves the average performance compared to non-meta counterparts, but also shows a steady increase in plasticity (decrease in avg.\ SAUCE) over time, indicating \emph{learning-to-learn} behavior. We highlight that this applies to \emph{both backward and forward per-shot plasticity}, and thus cannot simply be attributed to the additional forward information from the look-ahead examples. We present additional results on SAUCE in Appendix \ref{apx:more-results}.
\section{Discussion}\label{sec:discuss}

Inspired by cued recall and definitions of plasticity in natural learners, we explore a new paradigm for evaluation in continual learning: evaluation with few-shot adaptation on all tasks at all times. We construct task- and domain-incremental sequences from popular datasets for visual continual learning, focusing especially on variants that are longer than typically studied in CL. Through a large-scale fine-grained evaluation, we find that few-shot adaptation shows trends that deviate from conventional wisdom. We observe that model accuracy on backward tasks (stability) recovers strongly with very few examples, even when trained without methods that explicitly preserve stability. On some sequences, we observe a significant improvement even with only 1 or 2 examples per class, indicating that forgetting here is (at least in part) due to retrieval issues and not a catastrophic loss of knowledge. This finding actively impacts \emph{method selection} for CL---rather than store and replay samples for every task during training (expensive, and may potentially even hurt stability \cite{mahdaviyeh2025replay}), a practitioner may be able to rely on few-shot test-time adaptation to only the task at hand.

We also find that the prevalent definition of plasticity---the average accuracy immediately after training---is not only inadequate as a natural analogue, but also fails to capture the differences in adaptation ability of continually trained models. We remedy these issues by studying plasticity as the few-shot forward transfer over the sequence of learning rather than as an aggregated statistic. We observe multiple phenomena, both unexpected (such as the gradual loss of plasticity in EWC \citep{kirkpatrick2017overcoming}) and expected (such as AGEM \citep{AGEM} benefiting from task independence), that do not emerge with current evaluation strategies. We highlight the lacunae in existing meta-CL methods, and develop a simple \emph{foresight meta-learning} method to optimize for plasticity over the task sequence. We also introduce a novel metric to fully characterize plasticity---\emph{per-shot plasticity}---show that it highlights \emph{learning-to-learn} behavior when present. Overall, our work presents a new lens with which to view CL evaluation that is more practically useful and better aligned with natural learning systems.

{\bf Limitations.} Evaluation at scale naturally limits the exhaustiveness of parameter settings that can be studied (e.g., meta-learning hyperparameters), and this is a limitation of our work. We show that leveraging future examples for \emph{foresight} meta-learning can improve performance across CL methods, but we do not delve into the mechanism that facilitates this improvement and \emph{learning-to-learn} behavior. Finally, while we focus on the visual classification domain here, we note that adaptation comes in many forms. In future work, we intend to study adaptation with \emph{in-context learning} \cite{brown2020language} as an efficient measure of plasticity in continually learning large-scale foundation models. We also aim to further explore strategies for meta-learning few shot adaptation abilities over the task sequence.

\section*{Acknowledgment}
This work is supported by funds provided by the National Science Foundation and by DoD OUSD (R\&E) under Cooperative Agreement PHY-2229929 (The NSF AI Institute for Artificial and Natural Intelligence). We thank them for their generous support. We are also grateful to Tom Zollo and other members of the Zemel Group for their thoughtful feedback that helped improve this work.

\bibliography{references}
\bibliographystyle{abbrv}


\clearpage

\appendix

\section{Additional Related Work}\label{apx:more-related}

\subsection{Continual Learning Strategies}

Existing methods for continual learning can generally be categorized into replay-based, regularization-based, architectural, and distillation strategies. Of these, replay-based methods are by far the most popular (and successful) in the literature \cite{bidaki2025online}. These methods store or generate samples from previously learned tasks in a memory buffer, and utilize them to mitigate catastrophic forgetting during future training.
Experience Replay \cite{lin1992self} is a simple but effective method that augments the training dataset with buffered samples from previously learned tasks. More complex variants attempt to maximize the diversity of sampling \citep{bang2021rainbow}, find the best representative samples \citep{aljundi2019onlinecontinuallearningmaximally, liu2020mnemonics}, or generate synthetic samples from past tasks \cite{shin2017continual}. Similarly, Distillation methods augment downstream task learning with an objective that aims to replicate previously optimized model outputs on memory samples from earlier tasks. Dark Experience Replay \cite{buzzega2020darkexperiencegeneralcontinual} and its variants (DER++, X-DER) store the logits of the model along with the training samples and add a distillation loss to match these logits in downstream training. Learning without Forgetting \cite{li2017learningforgetting} uses a similar approach, but learns a new classification head for every task. Regularization-based approaches aim to prevent catastrophic forgetting by constraining model updates during downstream task training. Elastic Weight Consolidation \cite{kirkpatrick2017overcoming} regularizes the diagonal of the Fisher Information Matrix between the model weights of the current task and every previously seen task to limit the magnitude of updates to weights that are highly associated with performance on prior tasks. GEM and A-GEM \cite{lopez2017gradient, AGEM} project the model's gradient on a downstream task into the null space of the gradient over memory samples to prevent an optimizer step that increases loss on previous tasks. Architecture-based methods such as LwF \cite{li2017learningforgetting} and D-MoLE \cite{ge2025dynamic} learn a new set of model parameters for each new task. 
For a more thorough review, we refer the reader to well-researched surveys \cite{wang2024comprehensive, bidaki2025online} on continual learning.

\section{Meta-Learning and Meta-CL}\label{apx:meta-learning-overview}

Meta-learning algorithms for few-shot learning ability aim to retain or improve a machine learning model's few-shot performance on a set of pretraining tasks after that model is trained on a downstream task, either by learning a model initialization or an update rule for further training. Currently, the most popular methods focus on learning a good initialization via a model-agnostic procedure such as MAML \cite{finn2017modelagnosticmetalearningfastadaptation} or Reptile \cite{nichol2018firstordermetalearningalgorithms}. These methods learn a model initialization by leveraging the model loss (and its gradient) on a few examples of each pretraining task. The MAML algorithm optimizes the following objective over pretraining tasks $\mathbb{T}=\{\mathcal{T}_1,\ldots,\mathcal{T}_s\}$:
\begin{equation}
    \hat\theta = \theta - \beta\nabla_\theta\sum_{\mathcal{T}_i\sim p(\mathcal{T})} \mathcal{L}_{i}(\mathcal{T}_i^\mathrm{val}, U_k(\theta, \mathcal{T}_i^\mathrm{train}))
\end{equation}
where $\theta$ are the learning model parameters, $\beta$ is the meta-learning rate, $\mathcal{T}_i^\mathrm{train},\mathcal{T}_i^\mathrm{val}$ represent the training and validation splits of a pretraining task $\mathcal{T}_i$ sampled from $p(\mathcal{T})$ with loss metric $\mathcal{L}_i$, and $U(\alpha,\mathcal{D})$ is a differentiable optimizer of learning model $\alpha$ on at most $k$ examples of distribution $\mathcal{D}$ (and a corresponding objective $\mathcal{L}$, omitted here for clarity).

Since MAML requires the computation of expensive second-order derivatives to optimize over any gradient-based optimizer $U$, first-order approximations are often employed in practice. Among these, First-Order MAML \citep{finn2017modelagnosticmetalearningfastadaptation} is identical to MAML, except that it discards any second-order terms in the computation of the meta-update. Reptile \cite{nichol2018firstordermetalearningalgorithms} uses a finite-difference approximation of the gradient update, leading to the simple model-merging strategy defined below.
\begin{equation}
    \hat\theta = \theta - \beta\sum_{\mathcal{T}_i\sim p(\mathcal{T})}\theta - U_k(\theta, \mathcal{T}_i^\mathrm{train})
\end{equation}
With a small meta learning rate $\beta\le\epsilon$, the update is in the $\epsilon$-neighborhood of the base parameters $\theta$, and finite difference is a good approximation of the gradient of the meta objective.

Continual learning is a special case of the multi-task learning where a learner encounters tasks sequentially over the learning period, and must simultaneously perform well on all encountered tasks. While standard meta-learning algorithms assume access to every pretraining task prior to learning, the Online Meta-Learning algorithm \cite{pmlr-v97-finn19a} assumes a continuous stream of samples with no task boundaries, with a `follow the meta-leader' (FTML) protocol. This  algorithm applies MAML updates using a replay buffer $\nu_t$ of past tasks at time $t$, and fine-tunes the model on a task once all its samples are accumulated.
\begin{equation}
    \hat\theta_t = \theta_t - \beta\sum_{\mathcal{T}_i\sim \nu_t}\mathcal{L}_{i}(\mathcal{T}_i^\mathrm{val}, U_k(\theta, \mathcal{T}_i^\mathrm{train}))
\end{equation}

In effect, the above algorithm simulates the task-incremental continual learning setting with a replay buffer. Existing works on continual meta-learning (meta-CL) can be divided into two strategies. The first strategy modifies the \emph{architecture} of the learner using meta-information across the task sequence. ANML \citep{beaulieu2020learning}
produces an element-wise gating that modulates a model's task response. OML \citep{javed2019meta} freezes representation layers after a meta-training phase, only updating later layers. The second strategy uses a \emph{meta-optimizer} to leverage cross-task information while updating the model on each task. MER \cite{riemer2019learninglearnforgettingmaximizing} makes Reptile \cite{nichol2018firstordermetalearningalgorithms} meta-updates on every task and buffer sample to align samples. Online Meta-Learning \cite{pmlr-v97-finn19a} and La-MAML \cite{gupta2020lamamllookaheadmetalearning} use a fast/slow weight strategy to meta-learn information from past tasks prior to learning the current task. VR-MCL \cite{wu2024meta} approximates the cross-task Hessian from a replay buffer to inform optimization.

\section{Experimental Details}\label{apx:details}

\subsection{Datasets}
We evaluate on five continual learning task sequences spanning image classification tasks. Table~\ref{tab:datasets} summarizes the key properties of each task sequence.
 
\begin{table}[h]
\centering
\caption{Dataset statistics used in our experiments.}
\label{tab:datasets}
\small
\begin{tabular}{lccccc}
\toprule
\textbf{Dataset} & \textbf{Tasks} & \textbf{Classes/Task}
                 & \textbf{Image Size} & \textbf{Backbone} & \textbf{CL Setting} \\
\midrule
{\seqm}      & 5  & 2  & $28{\times}28$ & MLP       & Task-IL  \\
{\seqc}      & 10 & 10 & $32{\times}32$ & ResNet-18 & Task-IL  \\
{\seqct}     & 20 & 5  & $32{\times}32$ & ResNet-18 & Task-IL  \\
{\structc}   & 20 & 5 & $32{\times}32$ & ResNet-18 & Domain-IL (low overlap)  \\
{\rotm}      & 20 & 10 & $28{\times}28$ & ResNet-18 & Domain-IL (high overlap) \\
\bottomrule
\end{tabular}
\end{table}

\subsection{Code and Computer Resources}
To ensure correctness, we experiment using the Mammoth library for visual CL \cite{boschini2022class,buzzega2020dark} as our experimental testbed, which provides us with default implementations for the CL methods we evaluate (ER, DER++, EWC, AGEM, MER, and SGD). We conduct all experiments using a shared cluster with multiple NVIDIA A6000 GPUs, and a shared server with NVIDIA A100 GPUs. Each experiment involves up to 1000 cumulative epochs of training and 2000 evaluatory forward passes. Each epoch/pass takes on the order of seconds or minutes with our resources.

\subsubsection{CL Methods}
For each method and dataset, we train a single model sequentially on all tasks. Depending on task settings, we start with a randomly initialized ResNet-18 or an MLP with 32 hidden units. At the end of training for each task $t$, we save a checkpoint for k-shot adaptation and evaluation. For non-meta CL methods, the model is never exposed to data from future tasks during training.

We train each CL method using hyperparameters reported in the original papers for each dataset, as reported in the table below.

\begin{tabular}{c|l}
    \toprule
    Method & Hyperparameters \\
    \midrule
    SGD & $\text{lr}{=}0.1$ \\
    ER & $\text{lr}{=}0.1$,$|\text{buffer}|=500$ \\
    DER++ & $\text{lr}{=}0.03$, $\alpha{=}0.3$, $\beta{=}0.5$,$|\text{buffer}|=500$ \\
    EWC & $\text{lr}{=}0.1$, $\lambda{=}10$, $\gamma{=}1.0$;  \\
    AGEM & $\text{lr}{=}0.03$, $|\text{buffer}|=500$ \\
    MER & $\text{lr}{=}0.1$, $\beta{=}0.01$, $\gamma{=}0.03$,$|\text{buffer}|=200$ \\
    \bottomrule
\end{tabular}

Following the mammoth defaults, MER trains with a batch size of 1. We train every other method on {\seqc}, {\seqct} and {\structc} with batch size 32; {\seqm} and {\rotm} with batch size 64. We train for 50 epochs on variants of CIFAR100, for 5 epochs on {\rotm} and for 1 epoch on {\seqm}.

\subsubsection{Foresight Meta-Learning}
We implement foresight meta-learning as a wrapper class that can be composed with any base CL method (SGD, ER, DER++, EWC, A-GEM). The wrapper inserts a meta-update step at the beginning of each task, then the base method's training logic runs afterward unchanged. For meta-SGD, the inner loop uses SGD with no weight decay. For other CL methods (meta-ER, meta-DER++, meta-EWC, meta-AGEM), the inner loop uses the base method's own training logic, including any replay or regularization.

We train with the following hyperparameters. Reptile uses meta step size $\beta=0.2$ on {\seqc}, $\beta=0.1$ on {\seqct}, and $\beta=0.01$ on the remaining datasets; MAML uses $\beta=0.001$. We use $k=5$ examples per class per lookahead task across all datasets, giving a total of $N=10$ foresight examples for {\seqm}, $N=25$ for {\seqct} and {\structc}, and $N=50$ for the remaining datasets, with $L=1$ for {\seqm} and $L=3$ for all other datasets.

\subsection{Evaluation}
After training per-task checkpoints, we evaluate each checkpoint on the held-out test sets of \emph{all} tasks, both past and future relative to that checkpoint. Before evaluating performance on task $i$, we perform $k$-shot adaptation using the checkpoint saved after task $t$. We randomly sample $k$ examples from the training set of task $i$ and take 10 gradient steps on those examples with the adaptation learning rate reported on Table~\ref{tab:hparams_adapt}. The adapted model is then evaluated on the full test set of task $i$.

For meta-CL methods, the lookahead samples used during training are excluded from the forward transfer evaluation, to avoid measuring performance on tasks the model has already been explicitly optimised for.

\begin{table}[h]
\centering
\caption{Evaluation adaptation learning rates per method and dataset.}
\label{tab:hparams_adapt}
\small
\begin{tabular}{lcccccccc}
\toprule
\textbf{Dataset} & \textbf{SGD} & \textbf{ER} & \textbf{DER++} & \textbf{EWC} & \textbf{A-GEM} & \textbf{LWF} & \textbf{MER} \\
\midrule
{\seqm}    & 0.1  & 0.1  & 0.03  & 0.1  & 0.03  & 0.03  & 0.1   \\
{\seqc}    & 0.1  & 0.1  & 0.1   & 0.1  & 0.1   & 0.1   & 0.1   \\
{\seqct}   & 0.2  & 0.1  & 0.03  & 0.2  & 0.03  & 0.03  & 0.1   \\
{\structc} & 0.2  & 0.2  & 0.2   & 0.2  & 0.2   & 0.2   & 0.2   \\
{\rotm}    & 0.05 & 0.1  & 0.003 & 0.1  & 0.003 & 0.003 & 0.003 \\
\bottomrule
\end{tabular}
\end{table}

\section{Additional Results}\label{apx:more-results}

\begin{figure}
    \centering
    \includegraphics[width=0.8\textwidth]{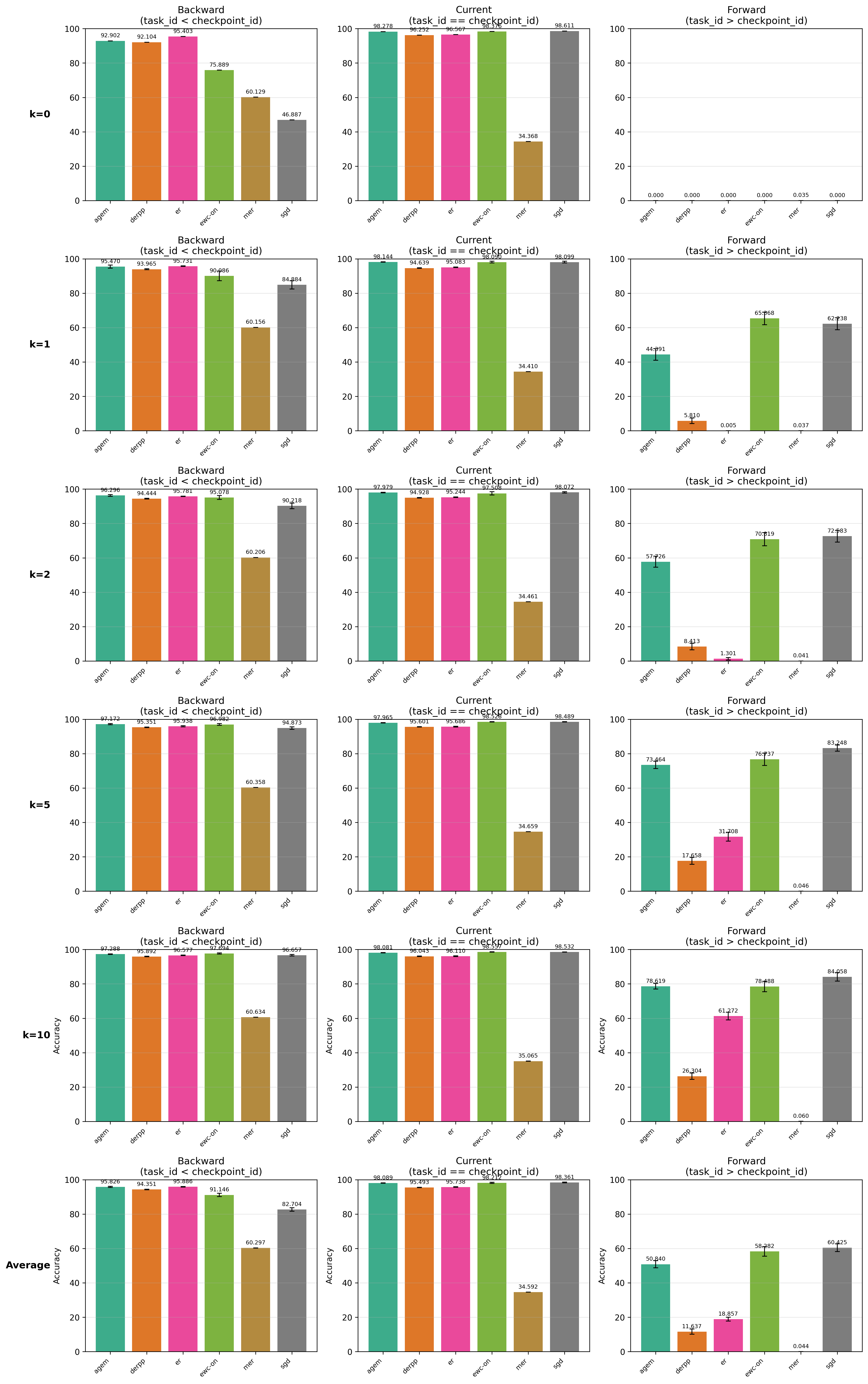}
    \caption{Mean $k$-shot accuracy of non-meta CL baselines on {\seqm} for
backward, current, and forward tasks, averaged across all checkpoints. Error
bars denote standard error over 10 seeds.}
    \label{fig:mnist-few-shot-full}
\end{figure}

\begin{figure}
    \centering
    \includegraphics[width=\textwidth]{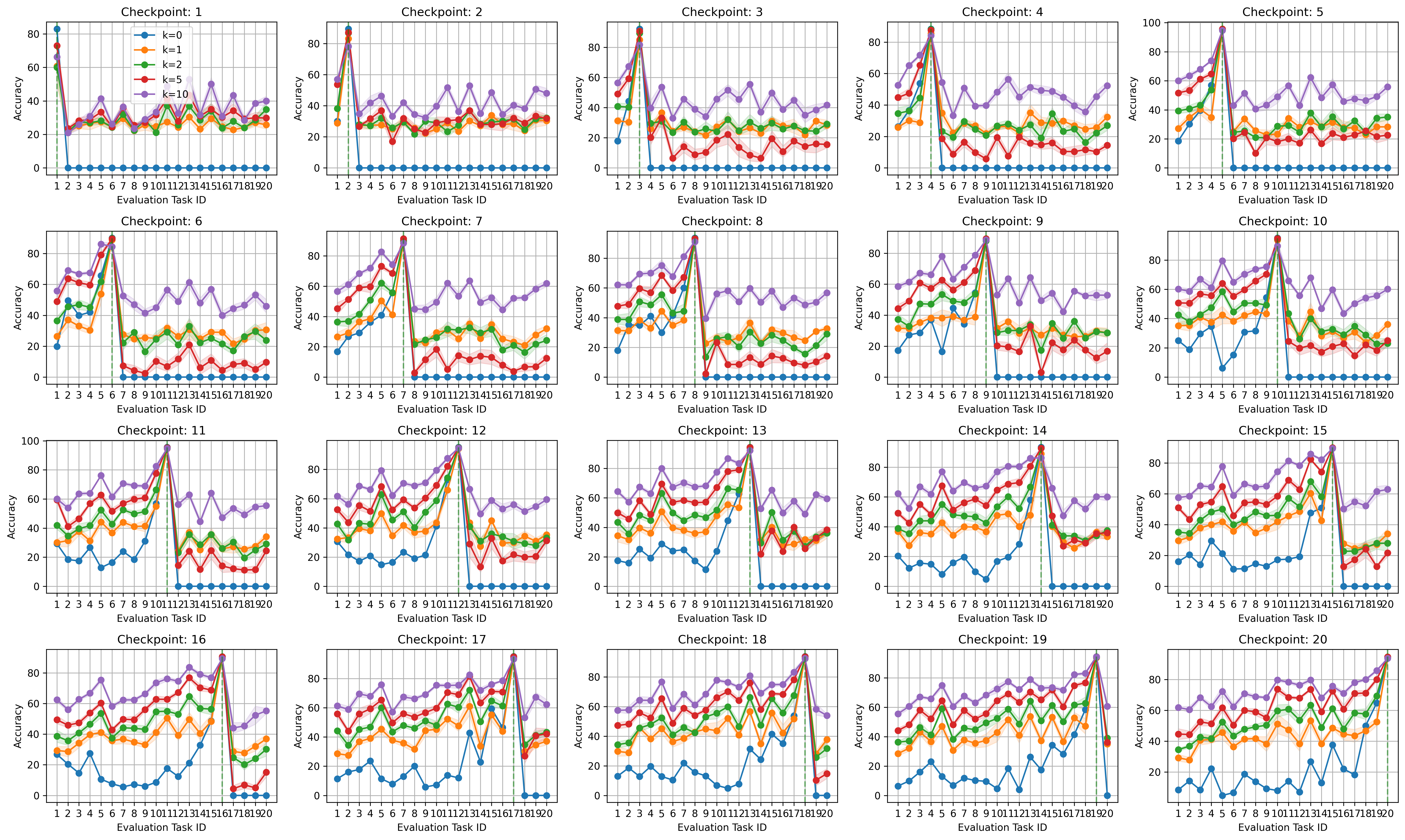}
    \caption{K-shot evaluation accuracy for SGD, evaluated on {\seqct} across 20 checkpoints. Each subplot shows accuracy on all 20 evaluation tasks at a given checkpoint (green dashed line). Lines represent different k-shot adaptation budgets ($k \in \{0, 1, 2, 5, 10\}$); shaded bands show standard error over 5 seeds.}
    \label{fig:accuracy-seq-c100}
\end{figure}

\begin{figure}
    \centering
    \includegraphics[width=\textwidth]{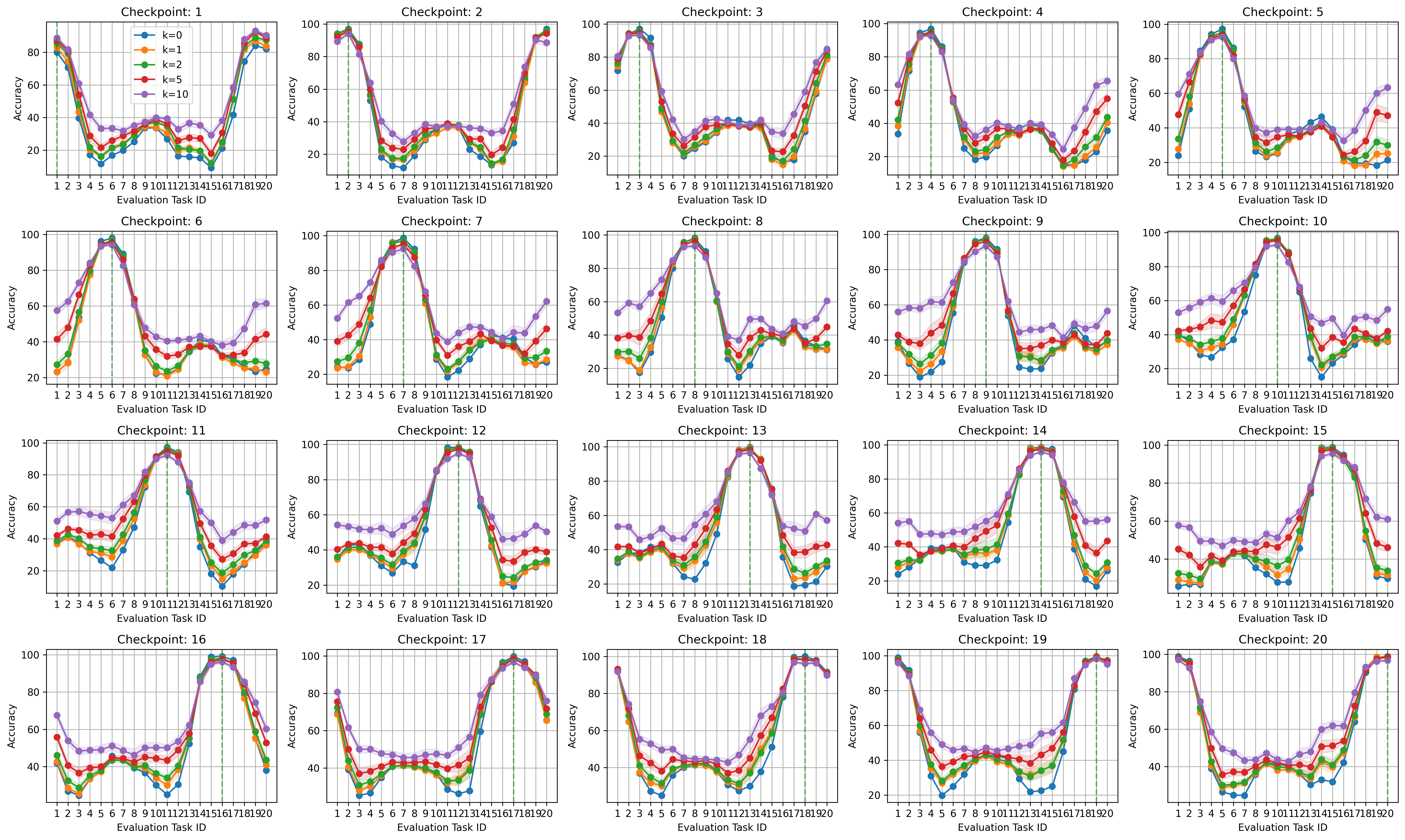}
    \caption{K-shot evaluation accuracy for SGD, evaluated on {\rotm} across 20 checkpoints. Each subplot shows accuracy on all 20 evaluation tasks at a given checkpoint (green dashed line). Lines represent different k-shot adaptation budgets ($k \in \{0, 1, 2, 5, 10\}$); shaded bands show standard error over 10 seeds.}
    \label{fig:accuracy-smooth-mnist}
\end{figure}

\begin{figure}
\centering
\begin{subfigure}{0.49\textwidth}
        \includegraphics[width=\textwidth]{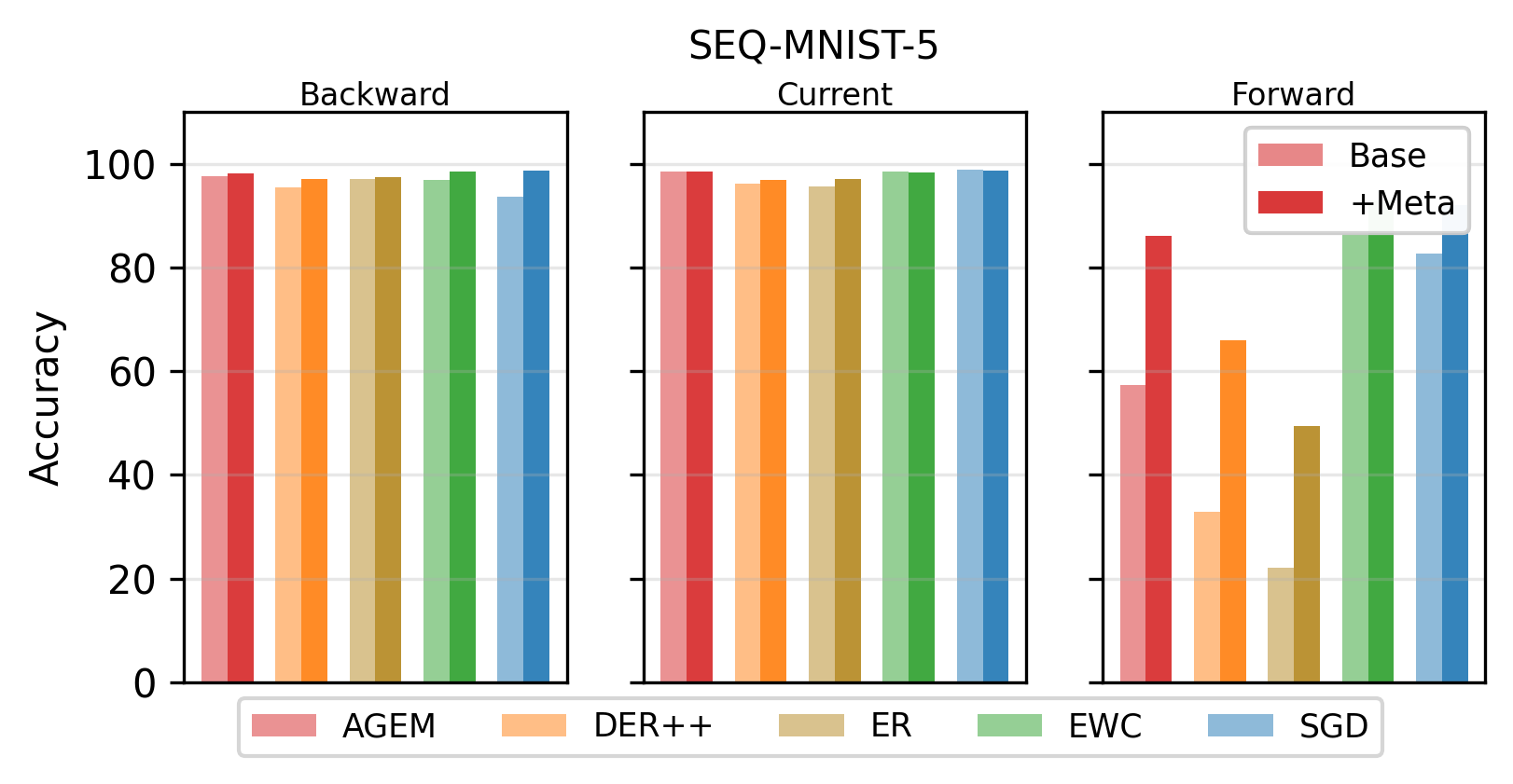}
    \label{fig:meta-cl-seq-mnist}
\end{subfigure}
\begin{subfigure}{0.49\textwidth}
        \includegraphics[width=\textwidth]{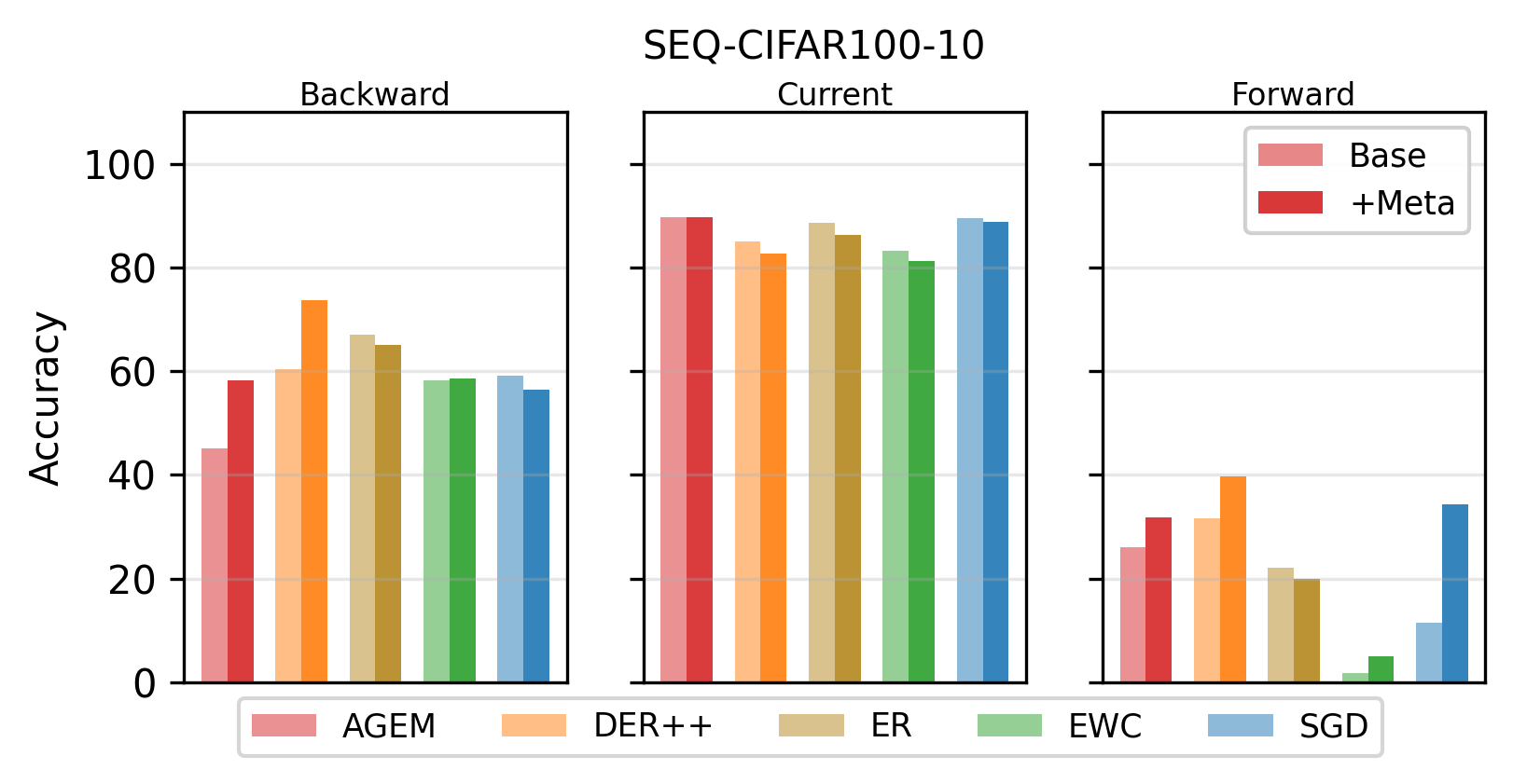}
    \label{fig:meta-cl-seq-cifar100}
\end{subfigure}
%
\begin{subfigure}{0.49\textwidth}
        \includegraphics[width=\textwidth]{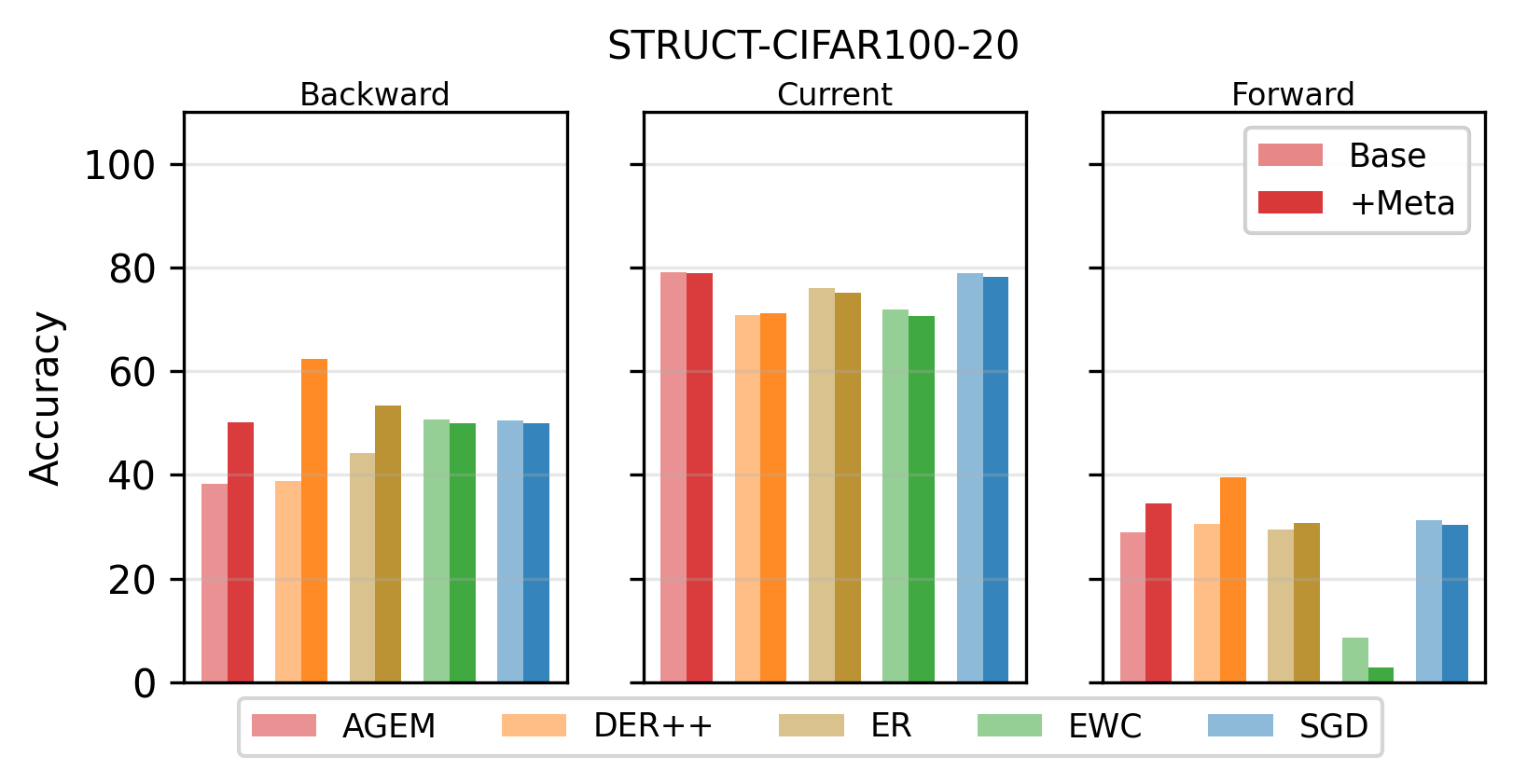}
    \label{fig:meta-cl-struct-cifar100}
\end{subfigure}
\begin{subfigure}{0.49\textwidth}
        \includegraphics[width=\textwidth]{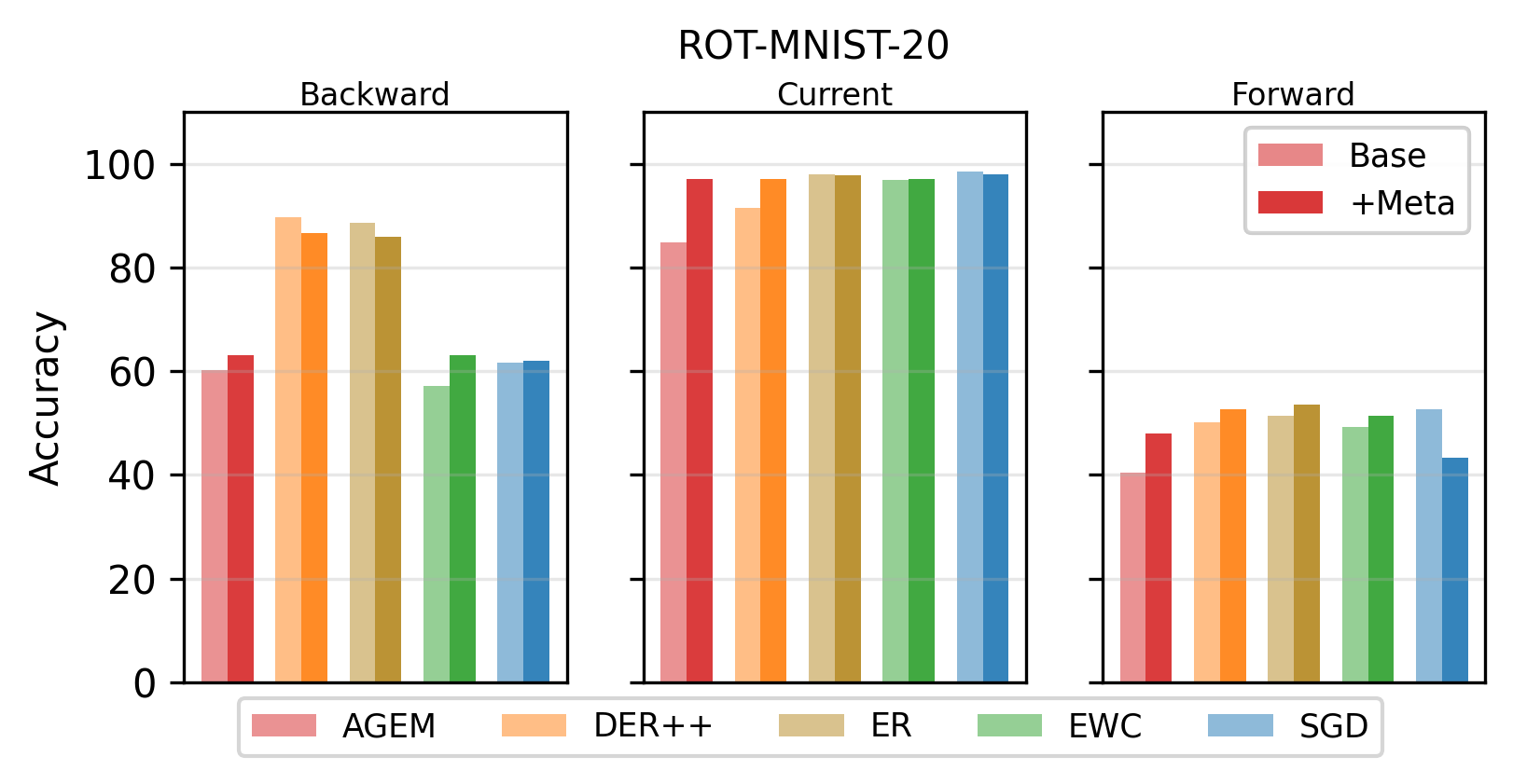}
    \label{fig:meta-cl-rot-mnist}
\end{subfigure}
\caption{Improvement in 5-shot forward transfer with \emph{foresight meta-learning} added to CL methods. We update each method with MAML on 1 ({\seqm}) or 3 (others) look-ahead tasks in parallel (Equation \ref{eqn:meta-parallel}). The forward measurement on meta methods \emph{does not} include look-ahead tasks.}
\label{fig:meta-cl-res-full}
\end{figure}

\begin{figure}
    \centering
    \includegraphics[width=0.8\textwidth]{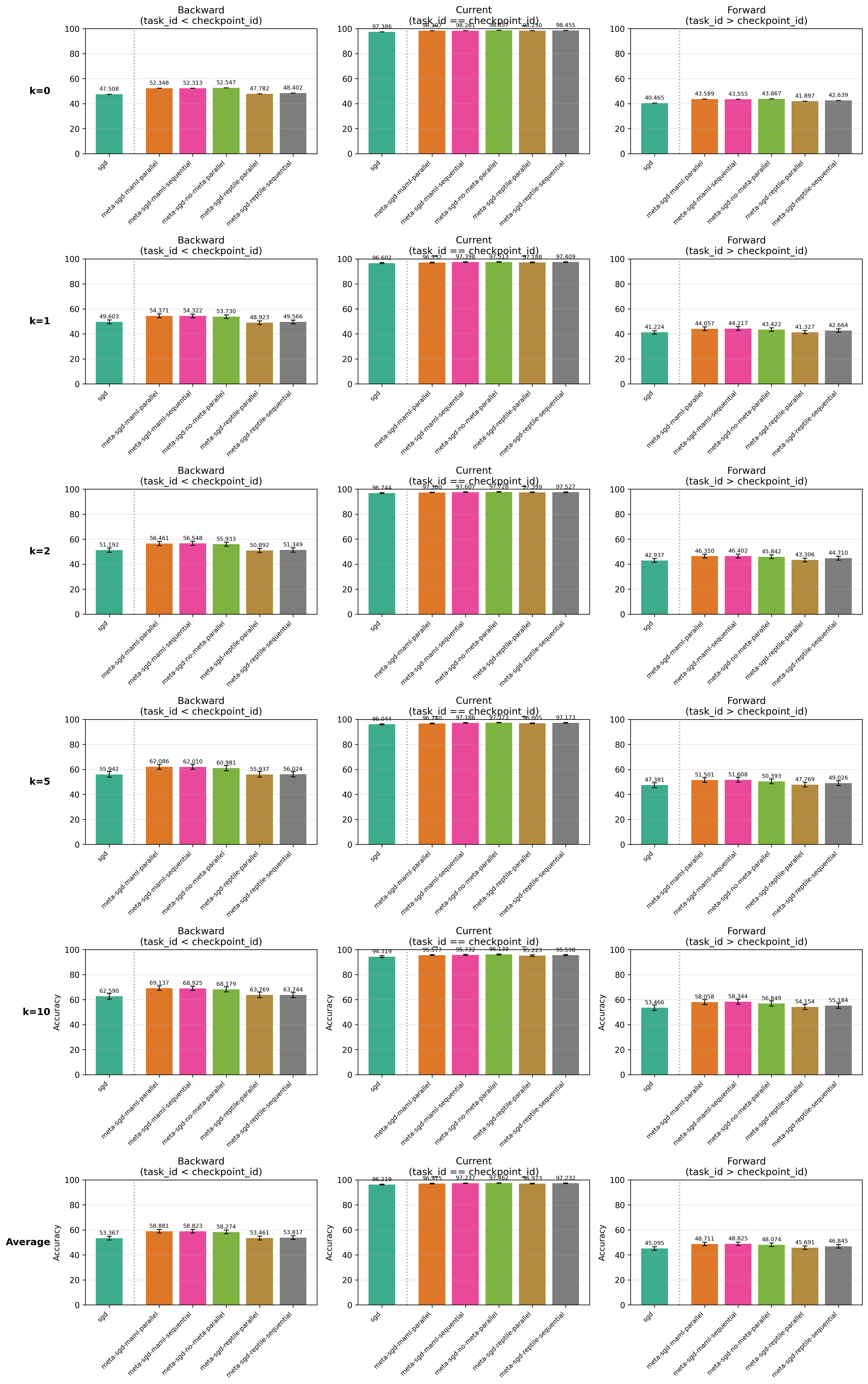}
    \caption{Mean $k$-shot accuracy of meta-SGD variants on {\rotm} for backward,current, and forward tasks, averaged across all checkpoints. Error bars denote standard error over 10 seeds.}
    \label{fig:smooth-mnist-meta-sgd-full}
\end{figure}

\begin{figure}
    \centering
    \includegraphics[width=\textwidth]{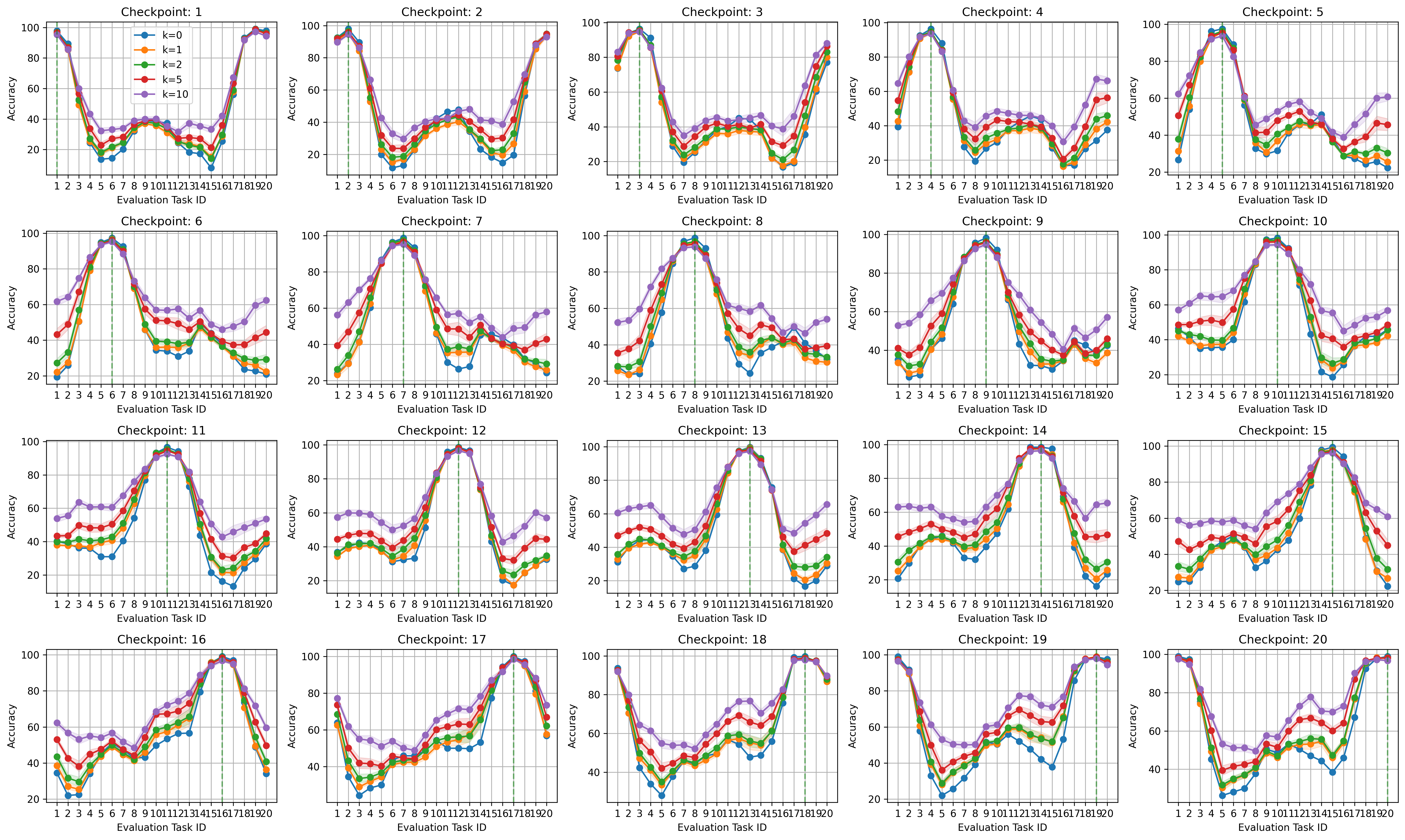}
    \caption{K-shot evaluation accuracy for parallel MAML with 20 adaptation steps and 20 meta-adaptation steps, evaluated on {\rotm} across 20 checkpoints. Each subplot shows accuracy on all 20 evaluation tasks at a given checkpoint (green dashed line). Lines represent different k-shot adaptation budgets ($k \in \{0, 1, 2, 5, 10\}$); shaded bands show standard error over 10 seeds.}
    \label{fig:smooth-mnist-maml-parallel-full}
\end{figure}

\begin{figure}
\centering
\begin{subfigure}{0.48\textwidth}
        \includegraphics[width=\textwidth]{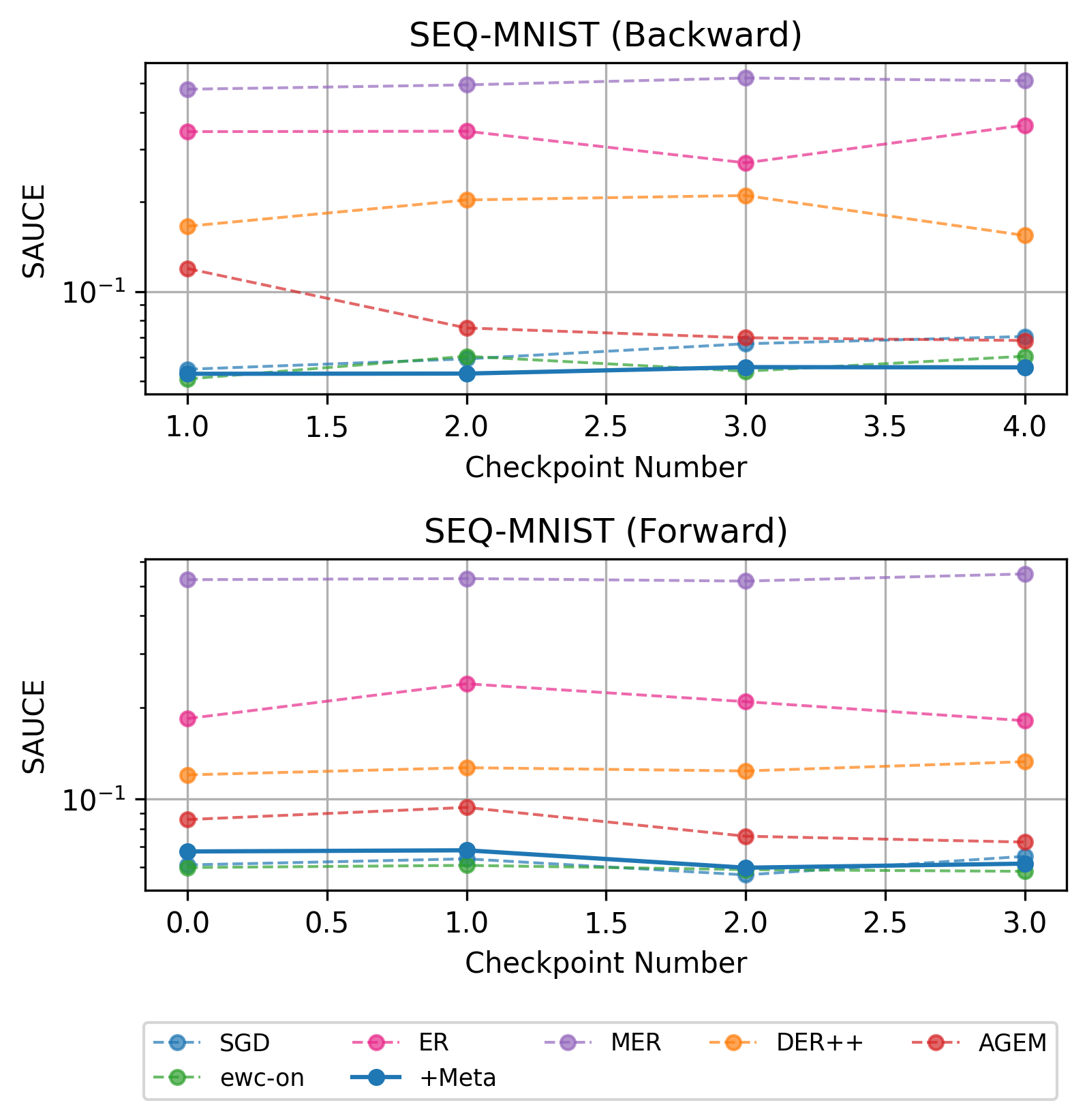}
    \label{fig:sauce-seq-mnist}
\end{subfigure}
%
%
\begin{subfigure}{0.48\textwidth}
        \includegraphics[width=\textwidth]{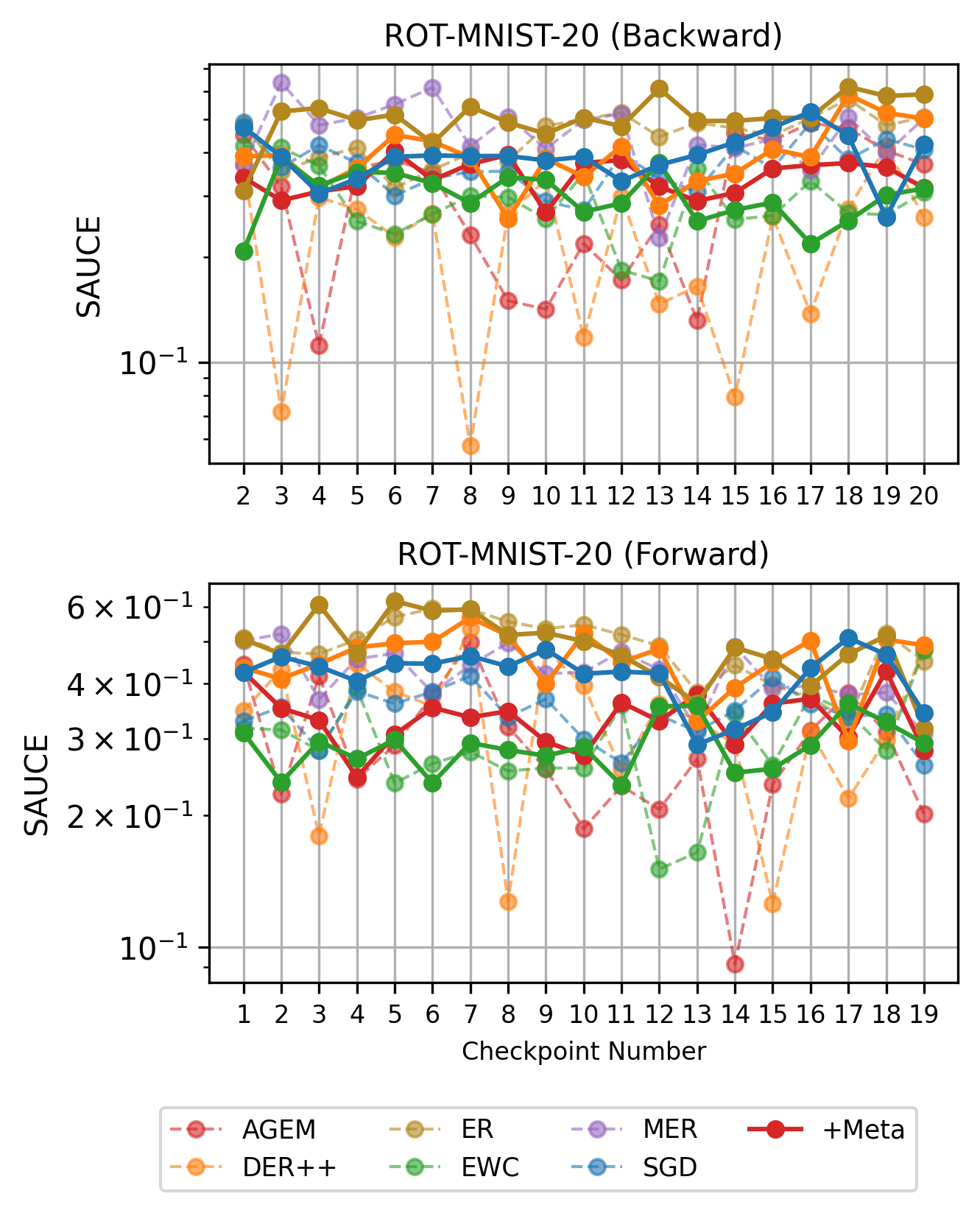}
    \label{fig:sauce-rot-mnist}
\end{subfigure}
\caption{Forward and backward per-shot plasticity on the task incremental {\seqm} and the domain-incremental {\rotm}. The solid lines represent methods augmented with \emph{foresight meta-learning}. The adaptation rate (SAUCE) does not show much variance over the short {\seqm} sequence, and fluctuates without much change on {\rotm}, even for \emph{foresight meta-learning} methods. We hypothesize that this is due to the nature of the distribution shift on {\rotm}. The entire input is rotated for every new task, but meta-learning methods appear to benefit from shared fine-grained structure across tasks (such as a common superclass).}
\label{fig:sauce-res-2}
\vspace{-0.5cm}
\end{figure}




\end{document}